%% file: neurips_2026.tex
\documentclass{article}

\PassOptionsToPackage{numbers,compress}{natbib}
	
	\usepackage[main, final]{neurips_2026}
	
	\usepackage[utf8]{inputenc} 
	\usepackage[T1]{fontenc}    
	\usepackage{hyperref}       
	\usepackage{url}            
	\usepackage{booktabs}       
	\usepackage{amsfonts}       
	\usepackage{nicefrac}       
	\usepackage{microtype}      
	\usepackage{xcolor}         
	
	\usepackage{amsmath, amssymb}
	\usepackage{amsthm}
	
	\usepackage{graphicx}
	\usepackage{caption}
	
	\usepackage{algorithm}
	\usepackage{algpseudocode}
	\title{The Cost of Mismatch: Noise Amplification in Zeroth-Order Reinforcement Learning}

	\author{%
		Lianmin Chen\textsuperscript{1} \quad Junbin Qiu\textsuperscript{2} \quad Chenxing Wei\textsuperscript{4} \quad	Yao Shu\textsuperscript{2,3}\thanks{Corresponding authors.} \quad Kun He\textsuperscript{1}\footnotemark[1] \\
		\textsuperscript{1}School of Computer Science and Technology, Huazhong University of Science and Technology \\
		\textsuperscript{2}The Hong Kong University of Science and Technology (Guangzhou) \\
		\textsuperscript{3}The Hong Kong University of Science and Technology\\
		\textsuperscript{4}The University of Hong Kong \\
		\texttt{lianminchen@hust.edu.cn, jqiu236@connect.hkust-gz.edu.cn}\\
		\texttt{weichenxing@connect.hku.hk, yaoshu@hkust-gz.edu.cn, brooklet60@hust.edu.cn}
	}
	
	\newtheorem{theorem}{Theorem}
	\newtheorem{lemma}{Lemma}
	\newtheorem{assumption}{Assumption}
	\newtheorem{corollary}{Corollary}
	\newtheorem{proposition}{Proposition}
	
\begin{document}

		\maketitle

		\begin{abstract}
			Zeroth-order (ZO) optimization has emerged as a promising approach for memory-efficient large language model (LLM) post-training, yet its reliability in reinforcement learning with verifiable rewards (RLVR) remains largely unexplored. Motivated by the distinct data regimes of on-policy RL and fixed-data optimization, we develop a generic variance-theoretic framework to investigate what limits the reliability of ZO gradient estimation. Our analysis reveals that estimation reliability is structurally determined by whether the two finite-difference evaluations are performed on \emph{matched data}. On-policy reinforcement learning (RL) under the mismatched-data regime produces a $1/\mu^2$ oracle-noise amplification that sets an irreducible noise floor as the perturbation scale $\mu$ decreases, whereas Boltzmann Targeted SFT (BOLT) under the matched-data regime eliminates this amplification via noise cancellation. We further derive convergence guarantees with explicit constants that cleanly separate the two regimes. These results motivate the ZO-BOLT algorithm, which builds on BOLT's matched-data advantage for reliable ZO optimization of an RLVR-aligned objective. Controlled perturbation-scale sweeps and matched-versus-mismatched end-to-end comparisons validate the predicted effect of data matching. Experiments across three models and two mathematical reasoning benchmarks further show that ZO-BOLT consistently improves over the pretrained baseline while retaining a near-inference memory footprint. The code is publicly available at \url{https://github.com/lianmin-chen/the_cost_of_mismatch}.
		\end{abstract}
		
		\section{Introduction}
		Reinforcement learning with verifiable rewards (RLVR) has become a key component in the post-training of large language models (LLMs)~\cite{shao2024deepseekmath,guo2025deepseek,wei2025redit,wang2026reschedule}. While first-order (FO) methods remain the default for this process, their reliance on backpropagation~\cite{rumelhart1986learning} imposes severe memory constraints due to the retention of intermediate activations, gradients, and often optimizer states~\cite{malladi2023fine, dang2026fzoo}.
		
		To mitigate these constraints, zeroth-order (ZO) optimization~\cite{spall1992multivariate, nesterov2017random} has received increasing attention as a compelling memory-efficient alternative. Instead of relying on backpropagation, ZO methods estimate gradients using only forward passes and thus reduce memory consumption to the inference level~\cite{malladi2023fine,zhang2024revisiting}. Prior studies have demonstrated the feasibility of memory-efficient LLM post-training with ZO optimization in supervised fine-tuning (SFT)~\cite{malladi2023fine,gautam2024variance,zhang2024revisiting,shu2026compander,qiu2026hessian}, and more recently, in preference optimization~\cite{galatolo2025visualising} (see App.~\ref{app:related_work} for related work). However, to the best of our knowledge, the reliability of ZO optimization for RLVR remains largely unexplored. This raises a key question: \emph{what limits ZO optimization from being reliably applied to RLVR in LLM post-training, and how can this limitation be addressed?}
		
		To answer this question, our key observation is that on-policy RL~\cite{williams1992simple, sutton1999policy, wang2022deep} introduces a different data regime from the fixed-dataset settings commonly considered in existing ZO post-training studies. Specifically, ZO methods typically estimate gradients via the finite difference of two function evaluations under a parameter perturbation. In fixed-dataset settings such as SFT, since data remains unchanged across parameter perturbations, the two function evaluations can be performed on the same minibatch, which we refer to as a \emph{matched-data} regime. In on-policy RL, however, each function evaluation uses data generated by the corresponding policy, so perturbing the parameters leads to different batches for the two evaluations, resulting in a \emph{mismatched-data} regime.
		
		In this work, we develop a generic variance-theoretic framework to characterize the reliability of ZO gradient estimation and analyze the impact of different data regimes. This framework decomposes the ZO gradient estimation error into a noise-free term and an oracle-noise term, revealing that the reliability of ZO gradient estimates is structurally determined by the noise correlation between the two finite-difference evaluations. In the \emph{mismatched-data} regime, instantiated by on-policy RL, function evaluations are performed on independently sampled data, leaving their stochastic fluctuations uncorrelated and inducing a $1/\mu^2$ amplification of oracle-noise variance, where $\mu$ denotes the perturbation scale. This establishes an irreducible noise floor as $\mu$ decreases, making it a key failure mechanism in direct ZO optimization for on-policy RL. Conversely, the \emph{matched-data} regime, instantiated by Boltzmann Targeted SFT (BOLT)~\cite{shu2026referencesampledboltzmannprojectionklregularized}, utilizes the same data across the finite-difference pair, enforcing aligned stochasticity and eliminating this amplification. Accordingly, these results identify BOLT as a natural basis for reliable ZO optimization, motivating ZO-BOLT as a practical matched-data ZO algorithm for optimizing the corresponding RLVR-aligned objective.
		
		Specifically, our contributions are:
		\begin{enumerate}
			\item \textbf{A generic variance-theoretic framework for noisy ZO optimization.} We develop a generic framework for analyzing ZO gradient estimation under noisy function evaluations. We establish Lem.~\ref{lem:noise_identity}, an exact decomposition of the ZO optimization estimator error, showing the oracle-noise term is entirely governed by the noise correlation between the two paired evaluations. This provides a unified foundation for analyzing different data regimes.
			
			\item \textbf{Variance-theoretic characterization of the mismatched- and matched-data regimes.} We instantiate the two data regimes with on-policy RL and Boltzmann Targeted SFT (BOLT), respectively. We prove that on-policy RL under the mismatched-data regime induces a $1/\mu^2$ oracle-noise amplification (Prop.~\ref{prop:online_rl_noise}), whereas BOLT under the matched-data regime eliminates this amplification, leaving only a bounded minibatch-level variance term (Prop.~\ref{prop:bolt_variance}). These results are unified into an oracle-noise separation theorem (Thm.~\ref{thm:oracle_separation}) and translated into convergence guarantees with explicit constants that cleanly separate the two regimes (Thm.~\ref{thm:zooconvergence}).
			
			\item \textbf{Empirical validation of noise amplification and practical evaluation of ZO-BOLT.} We first validate the predicted effect of the matched- and mismatched-data regimes on ZO optimization. Controlled perturbation-scale $\mu$ sweeps verify the $1/\mu^2$ amplification under independently sampled paired evaluations (Table~\ref{tab:controlled_noise} and Figure~\ref{fig:oracle_noise_scaling}), while matched-versus-mismatched end-to-end comparisons validate its optimization consequences (Figure~\ref{fig:matching_e2e} and Table~\ref{tab:matching_e2e}). We then evaluate ZO-BOLT across Qwen2.5-1.5B, Qwen2.5-7B, and DeepSeek-R1-Distill-Qwen-1.5B on GSM8K and MATH~\cite{qwen2025qwen25technicalreport,guo2025deepseek,cobbe2021gsm8k,hendrycks2021math}, showing consistent performance gains while retaining a near-inference memory footprint (Tables~\ref{tab:generalization} and~\ref{tab:memory_combined}).
		\end{enumerate}

		\section{Preliminaries}
		\label{sec:prelim}
		This section establishes the theoretical foundation for the analysis in Section~\ref{sec:ZOOnoise}. We begin by introducing the Zeroth-Order (ZO) gradient estimator with noisy function evaluations, which serves as the basic optimization primitive throughout this work. We then formalize the two primary settings considered in this work: on-policy Reinforcement Learning (RL) and Boltzmann Targeted SFT (BOLT). Although both target the same underlying objective, their reliance on online sampling versus offline fixed-data reuse gives rise to the mismatched- and matched-data regimes, respectively, leading to distinct noise dynamics in ZO gradient estimation (see Section~\ref{sec:ZOOnoise}).
		
		\subsection{ZO Gradient Estimator with Noisy Function Evaluations}
		We consider the minimization of an objective function $F : \mathbb{R}^d \to \mathbb{R}$ in settings where gradients are unavailable and only noisy function evaluations can be queried. In ZO optimization, gradients are approximated via finite differences of function values. Let $g = \nabla F(\theta)$ denote the gradient of $F$ at $\theta\in\mathbb{R}^d$. Following~\cite{nesterov2017random}, the standard forward finite-difference estimator $\hat g$ is defined as:
		\begin{equation}
			\label{eq:gradient_estimator}
			\hat g(\theta; u, \xi_+, \xi_-)
			=
			\frac{\tilde F(\theta + \mu u, \xi_+) - \tilde F(\theta, \xi_-)}{\mu}\, u,
			\quad
			u \sim \mathcal{N}(0, I_d),
			\quad
			\mu > 0.
		\end{equation}
		Here, $\mu$ denotes the perturbation scale, $u$ is a Gaussian search direction, and $\tilde F(\theta, \xi) = F(\theta) + e(\theta, \xi)$ is a noisy oracle evaluation. The random variable $\xi$ captures the source of stochasticity in the function evaluation (e.g., data sampling or environment noise), and $e(\theta, \xi)$ represents the resulting additive noise.
		
		\subsection{On-Policy Reinforcement Learning (RL)}
		
		We consider the standard on-policy RL objective~\cite{sutton1998reinforcement}
		\(
		J(\theta) = \mathbb{E}_{\tau \sim P_\theta}\bigl[\ell(\theta; \tau)\bigr]
		\)
		under a minimization convention, where $P_\theta$ denotes the trajectory distribution induced by policy $\pi_\theta$, and $\ell(\theta; \tau)$ denotes a scalar loss associated with trajectory $\tau$. A natural Monte Carlo oracle estimates $J(\theta)$ using $N$ independent rollouts~\cite{williams1992simple}:
		\begin{equation}
			\label{eq:mc_oracle}
			\tilde{J}(\theta, \xi)
			=
			\frac{1}{N} \sum_{n=1}^{N} \ell(\theta; \tau_n),
			\quad
			\tau_n \stackrel{\mathrm{i.i.d.}}{\sim} P_\theta.
		\end{equation}
		Critically, the trajectory distribution $P_\theta$ depends on the current 
		parameter $\theta$, so each oracle query requires fresh online rollouts from the policy being evaluated.
		
		\subsection{Boltzmann Targeted SFT}
		In contrast to on-policy RL, Boltzmann Targeted SFT (BOLT)~\cite{shu2026referencesampledboltzmannprojectionklregularized} operates in an offline fixed-dataset setting. Concretely, by reformulating the KL-regularized RLVR objective as an offline 
		importance-weighted supervised learning problem, BOLT decouples 
		data generation from policy optimization: trajectories are first collected 
		under a reference policy to construct a weighted dataset 
		\(\mathcal{D} = \{(x_i, y_i, w_i)\}_{i=1}^M\), where \(x_i\) is a prompt, \(y_i\) is a response sampled from the reference policy for \(x_i\),
		and each weight \(w_i > 0\) is fixed at construction time and independent of \(\theta\). Optimization then proceeds entirely over this fixed dataset via the following objective:
		\begin{equation}
			\label{eq:bolt_objective}
			F_{\mathrm{BOLT}}(\theta)
			=
			\frac{1}{M} \sum_{i=1}^{M} \ell_i(\theta),
			\quad
			\ell_i(\theta) = -w_i \log \pi_\theta(y_i \mid x_i).
		\end{equation}
		
		\section{Matched and Mismatched Data Regimes in Zeroth-Order Optimization}
		\label{sec:ZOOnoise}
		Building on Section~\ref{sec:prelim}, this section develops a precise variance-theoretic characterization of ZO gradient estimation. We first establish a generic variance-theoretic framework decomposing the noisy ZO estimator error into a noise-free and an oracle-noise term (Section~\ref{sec:framework}). We then show that on-policy RL under the mismatched-data regime has a $1/\mu^2$ amplification of the oracle-noise term (Section~\ref{sec:online}), while BOLT under the matched-data regime eliminates this effect (Section~\ref{sec:bolt}). These results are unified into an oracle-noise separation theorem (Section~\ref{sec:separation}) and translated into smoothed-objective stationarity guarantees with explicit constants (Section~\ref{sec:stationarity}).
		
		\subsection{A Variance-Theoretic Framework for Noisy ZO Gradient Estimation}
		\label{sec:framework}
		We begin by decomposing the ZO gradient estimator~\eqref{eq:gradient_estimator} into a noise-free term and an oracle-noise term to isolate the contribution of function-evaluation noise. We adopt the following standard assumption on the oracle.
		\begin{assumption}[Conditionally unbiased, square-integrable oracle]
			\label{ass:oracle}
			For every queried point $x\in\mathbb R^d$, the oracle returns $\tilde F(x,\xi)=F(x)+e(x,\xi)$, where $\mathbb E[e(x,\xi)\mid x]=0$ and $\mathbb E[e(x,\xi)^2\mid x]<\infty$.
		\end{assumption}
		Define the noise-free estimator $g_F(\theta;u):=[F(\theta+\mu u)-F(\theta)]u/\mu$. Substituting $\tilde{F}(x,\xi) = F(x) + e(x,\xi)$ into the noisy ZO gradient estimator~\eqref{eq:gradient_estimator} gives the exact decomposition:
		\begin{equation}
			\label{eq:zoo_decomposition}
			\hat{g}(\theta; u, \xi_+, \xi_-)
			=
			\underbrace{g_F(\theta; u)}_{\text{noise-free term}}
			+
			\underbrace{
				\frac{e(\theta + \mu u, \xi_+) - e(\theta, \xi_-)}{\mu}\, u
			}_{\text{oracle-noise term}}.
		\end{equation}
		
		The noise-free term $g_F$ corresponds to the standard Gaussian-smoothing gradient estimator whose bias and variance properties are well characterized in \cite{nesterov2017random}. The oracle-noise term captures the propagation of oracle noise $e(\theta, \xi)$ through the two function evaluations. Under Assumption~\ref{ass:oracle}, this term has zero mean, and its variance equals its second moment, which is governed by the dependence structure between $\xi_+$ and $\xi_-$. The following Lemma~\ref{lem:noise_identity} (proof in App.~\ref{app:proof_noise_identity}) formalizes this dependence.

		\begin{lemma}[Exact oracle-noise identity]
			\label{lem:noise_identity}
			Assume \(
			\mathbb{E}\!\left[
			\bigl(e(\theta+\mu u,\xi_+) - e(\theta,\xi_-)\bigr)^2 \,\|u\|^2
			\,\middle|\, \theta
			\right] < \infty.
			\)
			Then the following identity holds
			\begin{equation}
				\label{eq:noise_identity}
				\mathbb{E}\!\left[\|\hat g - g_F\|^2 \,\middle|\, \theta\right]
				=
				\frac{1}{\mu^2}\,
				\mathbb{E}\!\left[
				\bigl(e(\theta+\mu u,\xi_+) - e(\theta,\xi_-)\bigr)^2
				\,\|u\|^2
				\,\middle|\, \theta
				\right].
			\end{equation}
			
			Furthermore, if there exists $\nu^2 > 0$ such that for all $u$,
			\(
			\mathbb{E}\!\left[
			\bigl(e(\theta+\mu u,\xi_+) - e(\theta,\xi_-)\bigr)^2
			\,\middle|\, \theta, u
			\right]
			\le \nu^2,
			\)
			then $\mathbb E[\|\hat g-g_F\|^2\mid\theta]\le d\nu^2/\mu^2$,
			where $d$ arises from $\mathbb{E}[\|u\|^2] = d$ for $u \sim \mathcal{N}(0, I_d)$.
		\end{lemma}
		
		\paragraph{Remark.}
		Lem.~\ref{lem:noise_identity} explicitly isolates the contribution of oracle noise to the estimator error, and reveals that the ZO gradient estimator amplifies not the oracle noise itself, but the \emph{mismatch} between two noisy function evaluations. Consequently, the following analysis reduces to characterizing the conditional second moment
		\(
		\mathbb{E}\!\left[
		\bigl(e(\theta+\mu u,\xi_+) - e(\theta,\xi_-)\bigr)^2
		\,\middle|\, \theta, u
		\right]
		\)
		under different data regimes.
		
		A key implication arises when $\xi_+$ and $\xi_-$ are conditionally independent, in which the noise terms add rather than cancel; this is formalized in the following Corollary~\ref{cor:independent_noise} (proof in App.~\ref{app:proof_independent_noise}).
		
		\begin{corollary}[Independent evaluation noise]
			\label{cor:independent_noise}
			If $\xi_+$ and $\xi_-$ are conditionally independent given $(\theta,u)$, and
			\(
			\operatorname{Var}(e(\theta',\xi)\mid \theta') \le \sigma^2
			\quad \text{for all } \theta' \in \mathbb{R}^d,
			\)
			then
			\begin{equation}
				\label{eq:independent_noise}
				\mathbb{E}\!\left[\|\hat g - g_F\|^2 \,\middle|\, \theta\right]
				\le
				\frac{2\,d\,\sigma^2}{\mu^2}.
			\end{equation}
		\end{corollary}
		
		Cor.~\ref{cor:independent_noise} reveals that the $1/\mu^2$ amplification arises directly from evaluation independence. Whether this independence is structurally unavoidable depends on the data regime, which we analyze next.
		
		\subsection{On-policy RL: The Mismatched-data Regime}
		\label{sec:online}
		We now show that on-policy RL naturally induces independent evaluations, thereby instantiating the mismatched-data regime and making the $1/\mu^2$ amplification identified in Cor.~\ref{cor:independent_noise} unavoidable.
		
		\subsubsection{Mismatched Evaluations and Noise Amplification}
		
		Applying ZO optimization to the on-policy RL objective
		$J(\theta) = \mathbb{E}_{\tau \sim P_\theta}[\ell(\theta; \tau)]$
		yields the estimator:
		\begin{equation}
			\label{eq:zoo_online}
			\hat g_{\mathrm{on\_policy}}(\theta)
			=
			\frac{\tilde J(\theta + \mu u, \xi_+) - \tilde J(\theta, \xi_-)}{\mu}\, u,
			\quad g_J(\theta;u):=\frac{J(\theta+\mu u)-J(\theta)}{\mu}\,u.
		\end{equation}
		where $\tilde{J}$ denotes the Monte Carlo oracle defined in~\eqref{eq:mc_oracle}.
		
		We formalize the oracle behavior through the following assumption.
		\begin{assumption}[On-policy rollout oracle]
			\label{ass:online_oracle}
			For every queried point $x \in \mathbb{R}^d$, $
				\mathbb{E}[\tilde J(x,\xi) \mid x] = J(x)$ and $
				\operatorname{Var}(\tilde J(x,\xi) \mid x) \le \sigma_{\mathrm{roll}}^2$.
			Moreover, evaluations at $x_+ = \theta + \mu u$ and $x_- = \theta$ are conditionally independent given $(\theta,u)$.
		\end{assumption}
		
		This setting naturally instantiates the \textbf{mismatched-data regime}. Specifically, the evaluation at $\theta$ uses online rollouts sampled from $P_\theta$, while the evaluation at $\theta+\mu u$ uses online rollouts sampled from $P_{\theta+\mu u}$. These are different data-generating distributions, and for discrete autoregressive generation there is in general no construction to reuse identical trajectories across parameter values without violating the correct on-policy marginal distributions. Under this mismatched-data regime, the independent evaluations lead to $1/\mu^2$ \textbf{oracle-noise amplification}, as quantified by the following Proposition~\ref{prop:online_rl_noise} (proof in App.~\ref{app:proof_online_rl_noise}).
		
		\begin{proposition}[On-policy RL oracle-noise amplification]
			\label{prop:online_rl_noise}
			Under Assumptions~\ref{ass:oracle} and \ref{ass:online_oracle},
			\begin{equation}
				\label{eq:online_rl_bound}
				\mathbb{E}\!\left[\|\hat g_{\mathrm{on\_policy}} - g_J\|^2 \,\middle|\, \theta\right]
				\le
				\frac{2\,d\,\sigma_{\mathrm{roll}}^2}{\mu^2},
			\end{equation}
		\end{proposition}
		
		\paragraph{Severity of the bound.}
		The bound $2d\sigma_{\mathrm{roll}}^2/\mu^2$ exhibits double amplification: linear in dimension $d$ and quadratic in $1/\mu$. Even under a parameter-efficient configuration with $d=10^6$, the bound reaches $2\times10^{12}\sigma_{\mathrm{roll}}^2$ at $\mu=10^{-3}$, rendering the oracle-noise contribution prohibitively large and dominating any learning-rate-based mitigation. Crucially, this extreme magnitude is intrinsic to the ZO optimization framework rather than an artifact of a loose upper bound: the factor $d$ comes from the exact identity $\mathbb{E}[\|u\|^2]=d$, and $1/\mu^2$ is intrinsic to finite-difference estimation.
		
		\subsubsection{Limitations of Variance Reduction for ZO Optimization in On-Policy RL}
		
		\paragraph{Trajectory reuse changes the true objective.}
		A natural question is whether reusing the same trajectories across 
		evaluations can eliminate the $1/\mu^2$ amplification. However, our analysis (see App.~\ref{app:rl:reuse}) reveals that this is fundamentally limited: any variance reduction from trajectory reuse comes at the cost of replacing the true on-policy objective $J$ with a frozen-distribution surrogate $\bar{J}_\theta$, and the resulting distribution-shift term vanishes only when $P_{\theta+\mu u} = P_\theta$. Thus, any variance reduction from trajectory reuse is obtained by changing the optimization target.
		
		\paragraph{Importance-weighted correction.}
		The standard remedy is importance weighting, which corrects the frozen-distribution surrogate back to the true objective. However, the second moment of the required trajectory-level weights can grow exponentially with sequence length $T$, making unbiased correction unstable for long-form reasoning (see App.~\ref{app:rl:importance}).
		
		\paragraph{PPO/GRPO-style frozen-distribution surrogates.}
		Methods such as PPO~\cite{schulman2017proximal} and GRPO~\cite{shao2024deepseekmath} optimize clipped or locally constrained surrogates on rollouts from a frozen behavior policy. This permits paired ZO evaluations to reuse identical trajectories and yields the matched-data regime; their token-level or clipped ratios also depart from the exact trajectory-level importance correction to which the exponential second-moment result applies (see App.~\ref{app:rl:ppo}). However, these gains arise from replacing the true on-policy objective with a surrogate, rather than resolving the structural mismatch. Consequently, the $1/\mu^2$ amplification remains applicable to direct ZO optimization of the true on-policy RL objective.
		
		\subsection{BOLT: The Matched-data Regime}
		\label{sec:bolt}
		The preceding analysis shows that direct ZO optimization of on-policy RL incurs mismatch-induced $1/\mu^2$ oracle-noise amplification, while trajectory reuse avoids this mismatch only by changing the optimization target. We now turn to Boltzmann Targeted SFT (BOLT), introduced in Section~\ref{sec:prelim}, and establish that it structurally resolves the
		$1/\mu^2$ oracle-noise amplification identified in Section~\ref{sec:online}.
		
		\subsubsection{Matched Evaluations and Noise Cancellation}
		The key structural property of BOLT is that optimization operates over a fixed offline dataset, which allows the two function evaluations to use the \emph{same minibatch} $\mathcal{B}$, resulting in the \textbf{matched-data regime}. Specifically, for a minibatch $\mathcal{B} \subseteq \{1,\dots,M\}$ of size $B$, the ZO gradient estimator takes the form
		\begin{equation}
			\label{eq:zoo_bolt}
			\hat{g}_{\mathrm{BOLT}}(\theta; u, \mathcal{B})
			=
			\frac{F_{\mathcal{B}}(\theta + \mu u) - F_{\mathcal{B}}(\theta)}{\mu}\, u,
			\quad
			F_{\mathcal{B}}(\theta)
			=
			\frac{1}{B}\sum_{i \in \mathcal{B}} \ell_i(\theta).
		\end{equation}
		Here, $\ell_i(\theta)$ is defined in~\eqref{eq:bolt_objective}.

		This shared-data structure contrasts sharply with the on-policy online RL setting: the two evaluations are no longer driven by independent noise sources, but by a shared source of randomness. As we show in Prop.~\ref{prop:bolt_variance}, this difference fundamentally alters the scaling of the estimator variance.

		\subsubsection{Variance Bound with Explicit Constants}
		
		We first impose standard regularity conditions on the individual loss functions to facilitate the subsequent analysis.
		
		\begin{assumption}[Individual smoothness]
			\label{ass:smoothness}
			Each $\ell_i$ is $L$-smooth:
			\(
			\|\nabla \ell_i(\theta) - \nabla \ell_i(\theta')\|
			\le
			L \|\theta - \theta'\|.
			\)
		\end{assumption}
		
		\begin{assumption}[Uniform minibatch gradient variance]
			\label{ass:minibatch_variance}
			For every $x \in \mathbb{R}^d$,
			\begin{equation*}
				\mathbb{E}_{\mathcal{B}}
				\bigl[
				\|\nabla F_{\mathcal{B}}(x) - \nabla F_{\mathrm{BOLT}}(x)\|^2
				\bigr]
				\le
				\frac{\sigma_{\mathrm{mb}}^2}{B}.
			\end{equation*}
		\end{assumption}
		
		\paragraph{Remark.}
		Assumption~\ref{ass:smoothness} is standard in the nonconvex optimization literature. Assumption~\ref{ass:minibatch_variance} follows from the classical bounded-variance-per-sample condition
		\(
		\frac{1}{M}\sum_{i=1}^M \|\nabla \ell_i(\theta) - \nabla F_{\mathrm{BOLT}}(\theta)\|^2 \le \sigma_{\mathrm{mb}}^2
		\)
		via the standard minibatch variance reduction argument.
		
		We now quantify the variance reduction for BOLT under the matched-data regime in Proposition~\ref{prop:bolt_variance} (proof in App.~\ref{app:proof_bolt}).
		
		\begin{proposition}[Oracle-noise bound for BOLT under the matched-data regime]
			\label{prop:bolt_variance}
			Under Assumptions~\ref{ass:smoothness} and~\ref{ass:minibatch_variance}, for every $\theta \in \mathbb{R}^d$,
			\begin{equation}
			\mathbb{E}\!\left[\|\hat g_{\mathrm{BOLT}} - g_{F_{\mathrm{BOLT}}}\|^2 \,\middle|\, \theta\right]
			\le
			\left(
			\sqrt{\frac{(d+2)\sigma_{\mathrm{mb}}^2}{B}}
			+
			L\mu\sqrt{d(d+2)(d+4)}
			\right)^2,
		\end{equation}
			where $g_{F_{\mathrm{BOLT}}}(\theta;u) := \frac{F_{\mathrm{BOLT}}(\theta+\mu u)-F_{\mathrm{BOLT}}(\theta)}{\mu}\,u$.
		\end{proposition}
		
		\paragraph{Absence of $1/\mu^2$ amplification.}
		\label{rem:bolt_no_amplification}
		In contrast to the on-policy RL setting, the variance bound under BOLT contains \emph{no} $1/\mu^2$ term. Because the same minibatch $\mathcal{B}$ is used in both $F_{\mathcal{B}}(\theta + \mu u)$ and $F_{\mathcal{B}}(\theta)$, their leading-order stochastic coefficients cancel in the finite-difference operation, entirely eliminating the catastrophic amplification identified in 
		Section~\ref{sec:online}. The resulting bound does contain an $\mathcal{O}(d^3)$ term, but this is negligible in practice under the standard condition $\mu^2 d^2 L^2 \ll \sigma_{\mathrm{mb}}^2/B$, ensuring that the minibatch-noise term dominates.
		
		\subsection{The Oracle-Noise Separation Theorem}
		\label{sec:separation}
		We now state our main theorem, which synthesizes the preceding analysis into a single quantitative contrast between mismatched-data and matched-data regimes.
		\begin{theorem}[Oracle-noise separation]
			\label{thm:oracle_separation}
			Under Assumptions~\ref{ass:oracle}--\ref{ass:minibatch_variance}, we distinguish two regimes: on-policy RL under the mismatched-data regime and BOLT under the matched-data regime.\\
			(i) on-policy RL (the mismatched-data regime).
			\begin{equation}
				\mathbb{E}\!\left[\|\hat g_{\mathrm{on\_policy}} - g_J\|^2 \,\middle|\, \theta\right]
				\le
				\frac{2\,d\,\sigma_{\mathrm{roll}}^2}{\mu^2}.
			\end{equation}
			
			(ii) BOLT (the matched-data regime).
			\begin{equation}
				\mathbb{E}\!\left[\|\hat g_{\mathrm{BOLT}} - g_{F_{\mathrm{BOLT}}}\|^2 \,\middle|\, \theta\right]
				\le
				\left(
				\sqrt{\frac{(d+2)\sigma_{\mathrm{mb}}^2}{B}}
				+
				L\mu\sqrt{d(d+2)(d+4)}
				\right)^2.
			\end{equation}
			The key distinction is that matched minibatches remove the $1/\mu^2$ amplification in the oracle-noise term $\hat g - g_F$. They do not eliminate the intrinsic random-direction variability of one-point zeroth-order estimation.
		\end{theorem}
		
		\subsection{Convergence to Approximate Stationarity}
		\label{sec:stationarity}
		Thm.~\ref{thm:oracle_separation} establishes a sharp oracle-noise separation between on-policy RL and BOLT. This separation directly enables convergence analysis under both settings. We now combine the variance bounds from Thm.~\ref{thm:oracle_separation} with a standard descent lemma for Gaussian-smoothed nonconvex optimization to derive rates of convergence to approximate stationarity.
		Assume the objective function $F$ (either $J$ in on-policy RL or $F_{\mathrm{BOLT}}$ in BOLT) is $L_F$-smooth, and define its Gaussian smoothing
		\(
		F_\mu(\theta):=\mathbb E_{u\sim\mathcal N(0,I_d)}[F(\theta+\mu u)]
		\). We further impose the following standard assumptions.
		
		\begin{assumption}[Smoothed initial suboptimality]
			\label{ass:initial_suboptimality}
			Let
			\(
			F^* := \inf_{\theta \in \mathbb{R}^d} F(\theta),
			\)
			and assume that
			\(
			F(\theta_0) - F^* \leq \Delta_0 < \infty.
			\)
			For the Gaussian-smoothed objective
			\(
			F_\mu(\theta),
			\)
			define
			\(
			\Delta_{\mu,0}
			:=
			F_\mu(\theta_0)
			-
			\inf_{\theta \in \mathbb{R}^d}F_\mu(\theta).
			\)
			Under $L_F$-smoothness,
			\[
			\Delta_{\mu,0}
			\leq
			\Delta_0+\frac{L_F\mu^2d}{2}
			<
			\infty.
			\]
		\end{assumption}
		
		The following lemma (proof in App.~\ref{app:proof_smoothing}) collects standard properties of Gaussian smoothing.
		\begin{lemma}[Gaussian smoothing identity and gradient comparison]
			\label{lem:smoothing}
			\(
			\mathbb{E}_u[g_F(\theta;u)] = \nabla F_\mu(\theta),
			\)
			and
			\(
			\|\nabla F_\mu(\theta) - \nabla F(\theta)\|
			\le
			L_F \mu\,\mathbb{E}[\|u\|]
			\le
			L_F \mu \sqrt{d}.
			\)
		\end{lemma}
		
		Moreover, the second moment of the noise-free estimator satisfies Lemma~\ref{lem:zo_second_moment} (proof in App.~\ref{app:proof_zo_second_moment}):
		
		\begin{lemma}[Second moment of the noise-free one-point forward estimator]
			\label{lem:zo_second_moment}
			If $F$ is $L_F$-smooth, then
			\begin{equation}
				\mathbb{E}_u\!\left[\|g_F(\theta;u)\|^2\right]
				\le
				4(d+2)\|\nabla F_\mu(\theta)\|^2
				+
				\frac{L_F^2 \mu^2\,d(d+2)(d+12)}{2}.
			\end{equation}
		\end{lemma}
		
		We now state a unified convergence result for ZO methods in Theorem~\ref{thm:zooconvergence} (proof in App.~\ref{app:proof_zooconvergence}).
		\begin{theorem}[Smoothed nonconvex descent with explicit one-point ZO term]
			\label{thm:zooconvergence}
			Consider the update: 
			\(
			\theta_{t+1} = \theta_t - \eta \hat g_t, u_t \stackrel{\mathrm{i.i.d.}}{\sim} \mathcal{N}(0,I_d).
			\)
			Assume that for every $t$,
			\(
			\mathbb{E}[\hat g_t - g_F(\theta_t;u_t) \mid \theta_t, u_t] = 0, \mathbb{E}[\|\hat g_t - g_F(\theta_t;u_t)\|^2 \mid \theta_t] \le V,
			\)
			almost surely, where $V$ is a deterministic upper bound. If
			\(
			0 < \eta \le \frac{1}{4L_F(d+2)},
			\)
			then
			\[
			\frac{1}{T}\sum_{t=0}^{T-1}
			\mathbb{E}\!\left[\|\nabla F_\mu(\theta_t)\|^2\right]
			\le
			\frac{2\Delta_{\mu,0}}{\eta T}
			+
			L_F \eta
			\left(
			V + \frac{L_F^2 \mu^2\,d(d+2)(d+12)}{2}
			\right).
			\]
			
			In particular, combining with Theorem~\ref{thm:oracle_separation} gives:
			
			(i) on-policy RL (the mismatched-data regime).
			\[
			\frac{1}{T}\sum_{t=0}^{T-1}
			\mathbb{E}\!\left[\|\nabla J_\mu(\theta_t)\|^2\right]
			\le
			\frac{2\Delta_{\mu,0}}{\eta T}
			+
			L_J \eta
			\left(
			\frac{2d\sigma_{\mathrm{roll}}^2}{\mu^2}
			+
			\frac{L_J^2 \mu^2\,d(d+2)(d+12)}{2}
			\right).
			\]
			
			(ii) BOLT (the matched-data regime).
			\begin{align*}
				\frac{1}{T}\sum_{t=0}^{T-1}
				\mathbb{E}\!\left[\|\nabla (F_{\mathrm{BOLT}})_\mu(\theta_t)\|^2\right]
				&\le
				\frac{2\Delta_{\mu,0}}{\eta T}
				+
				L_{F_{\mathrm{BOLT}}} \eta
				\left[
				\left(
				\sqrt{\frac{(d+2)\sigma_{\mathrm{mb}}^2}{B}}
				+
				L\mu\sqrt{d(d+2)(d+4)}
				\right)^2
				\right.\\
				&\quad\left.
				+
				\frac{L_{F_{\mathrm{BOLT}}}^2 \mu^2\,d(d+2)(d+12)}{2}
				\right].
			\end{align*}
		\end{theorem}
		
		\paragraph{Remark.}
		Thm.~\ref{thm:zooconvergence} makes the cost of mismatched evaluations concrete. For on-policy RL under the mismatched-data regime, the $1/\mu^2$ amplification creates an irreducible bias-variance tension: shrinking $\mu$ to reduce smoothing bias simultaneously inflates the noise floor, imposing a hard lower bound on the achievable gradient norm that is independent of $T$. For BOLT under the matched-data regime, the absence of this factor means the bound is fully controllable via $(\mu, B, \eta, T)$, allowing convergence to any target accuracy $\epsilon$.
		
		\section{Experiments}
		\label{sec:experiments}
		In this section, we first validate the predicted effects of the matched- and mismatched-data regimes on ZO optimization through perturbation-scale $\mu$ sweeps and end-to-end comparisons (Section~\ref{sec:data_matching_validation}). Building on these findings, we evaluate ZO-BOLT, a practical matched-data ZO algorithm for optimizing the RLVR-aligned objective, in terms of performance and memory efficiency across different models and mathematical reasoning benchmarks (Section~\ref{sec:zobolt_effectiveness}). Experimental details are in App.~\ref{app:experiments}.
		
		\subsection{Controlled Comparison of the Matched- and Mismatched-Data Regimes}
		\label{sec:data_matching_validation}
\begin{figure}[h]
			\centering
			\begin{minipage}[c]{0.50\textwidth}
			\captionof{table}{\textbf{Dimension-normalized oracle-noise variance $V(\mu)$ across perturbation scales.} Matched BOLT shares the same minibatch; mismatched BOLT and on-policy RL sample data independently.}
			\label{tab:controlled_noise}
			\centering
				\vspace{4pt}
				\small
				\setlength{\tabcolsep}{1.5pt}
			\begin{tabular}{cccc}
				\toprule
				$\mu$ & $\textbf{BOLT}_\textbf{match}$ & $\textbf{BOLT}_\textbf{mismatch}$ & \textbf{RL} \\
				\midrule
				$10^{-4}$          & $\boldsymbol{2.12{\times}10^{-2}}$ & $2.77{\times}10^6$ & $2.44{\times}10^6$ \\
				$3{\times}10^{-4}$ & $\boldsymbol{2.12{\times}10^{-2}}$ & $3.07{\times}10^5$ & $2.63{\times}10^5$ \\
				$10^{-3}$          & $\boldsymbol{2.12{\times}10^{-2}}$ & $2.76{\times}10^4$ & $2.49{\times}10^4$ \\
				$3{\times}10^{-3}$ & $\boldsymbol{2.19{\times}10^{-2}}$ & $3.07{\times}10^3$ & $3.21{\times}10^3$ \\
				$10^{-2}$          & $\boldsymbol{3.31{\times}10^{-2}}$ & $2.77{\times}10^2$ & $2.30{\times}10^2$ \\
				\bottomrule
			\end{tabular}
				\vspace{4pt}
			\end{minipage}
			\hfill
			\begin{minipage}[c]{0.48\textwidth}
				\centering
				\includegraphics[width=\textwidth]{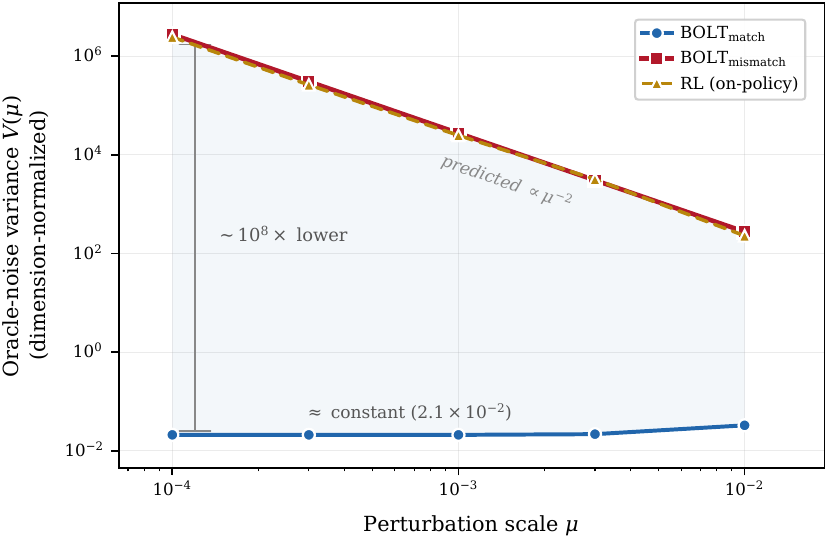}
				\caption{\textbf{Oracle-noise variance across perturbation scales.} Matched BOLT remains nearly constant, while mismatched BOLT and on-policy RL follow the predicted $1/\mu^2$ trend. Both axes are logarithmic.}
				\label{fig:oracle_noise_scaling}
			\end{minipage}
		\end{figure}

		\paragraph{Oracle-noise variance across perturbation scales.}
		Following our theory, we directly estimate the oracle-noise variance for on-policy RL under the mismatched-data regime and BOLT under the matched-data regime, as characterized by Prop.~\ref{prop:online_rl_noise} and Prop.~\ref{prop:bolt_variance}. For BOLT, an independent-minibatch variant, termed mismatched BOLT, is included to isolate the impact of data matching. Using Qwen2.5-1.5B with 1.09M trainable LoRA parameters on GSM8K, we sweep five perturbation scales and report the dimension-normalized oracle-noise variance $V(\mu)=d^{-1}\mathbb E\|\hat g-g_F\|^2$ in Table~\ref{tab:controlled_noise} and Figure~\ref{fig:oracle_noise_scaling}. Full experimental details are provided in App.~\ref{app:variance_experiments}.
		
		Table~\ref{tab:controlled_noise} and Figure~\ref{fig:oracle_noise_scaling} show that $V(\mu)$ grows approximately as $1/\mu^2$ as $\mu$ decreases for mismatched BOLT and on-policy RL, while remaining nearly constant for matched BOLT. These results confirm the predicted separation between the two data regimes in terms of oracle-noise amplification.

\begin{figure}[h]
			\centering
			\begin{minipage}[c]{0.53\textwidth}
				\centering
				\includegraphics[width=\textwidth]{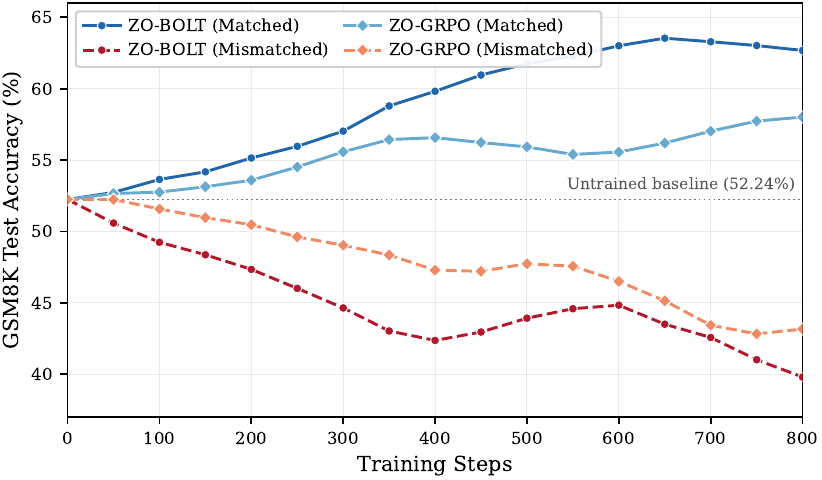}
				\caption{\textbf{GSM8K test accuracy vs.\ training steps.} Four-point smoothed curves end at Step~800; markers indicate evaluated checkpoints. The dashed line marks the pretrained baseline ($52.24\%$).}
				\label{fig:matching_e2e}
			\end{minipage}
			\hfill
			\begin{minipage}[c]{0.45\textwidth}
				\captionof{table}{\textbf{Best and final GSM8K test accuracy ($\%$)} in the matched-versus-mismatched control. $\Delta$ Best is the absolute gain over the pretrained baseline ($52.24\%$). Column-wise best and second-best results are bolded and underlined; gains are red.}
				\label{tab:matching_e2e}
				\centering
				\vspace{4pt}
				\small
				\setlength{\tabcolsep}{1.5pt}
				\begin{tabular}{llrrr}
					\toprule
					\textbf{Method} & \textbf{Data Regime} & \textbf{Best} & \textbf{$\Delta$ Best} & \textbf{Final} \\
					\midrule
					Baseline & --          & 52.24 & --        & 52.24 \\
					\textbf{ZO-BOLT} & Matched    & \textbf{63.78} & \textcolor{red!75!black}{$\boldsymbol{+11.54}$} & \textbf{62.95} \\
					ZO-BOLT & Mismatched & 49.35 & \textcolor{red!75!black}{$-2.89$}  & 41.76 \\
					ZO-GRPO & Matched    & \underline{59.00} & \textcolor{red!75!black}{$\underline{+6.76}$}  & \underline{57.56} \\
					ZO-GRPO & Mismatched & 53.08 & \textcolor{red!75!black}{$+0.84$}  & 44.80 \\
					\bottomrule
				\end{tabular}
				\vspace{4pt}
			\end{minipage}
		\end{figure}

		\paragraph{End-to-end performance.}
		To assess whether the estimator-level separation translates into end-to-end performance, we compare matched- and mismatched-data variants of ZO-BOLT and ZO-GRPO using Qwen2.5-1.5B with LoRA on GSM8K. The matched variants reuse the same minibatch across paired function evaluations, whereas the mismatched variants sample minibatches independently. While ZO-GRPO optimizes a frozen-distribution surrogate rather than the true on-policy objective, we include ZO-GRPO as the ZO counterpart of the widely used RLVR method GRPO, providing a practical setting for evaluating data-regime effects. Figure~\ref{fig:matching_e2e} and Table~\ref{tab:matching_e2e} report the results, with experimental details in App.~\ref{app:matching_optimization}.

		Figure~\ref{fig:matching_e2e} and Table~\ref{tab:matching_e2e} show that matched-data variants of both ZO-BOLT and ZO-GRPO consistently outperform their mismatched counterparts, with ZO-BOLT achieving larger gains over the pretrained baseline. Together with the oracle-noise measurements, these results demonstrate the impact of different data regimes on ZO optimization performance.
		
		\subsection{Effectiveness and Memory Efficiency of ZO-BOLT}
		\label{sec:zobolt_effectiveness}
		\label{sec:main_results}
		\label{sec:memory_efficiency}
		Having established the benefits of the matched-data regime, we further evaluate ZO-BOLT, a practical matched-data ZO algorithm for optimizing the corresponding RLVR-aligned objective, across different models and mathematical reasoning benchmarks, focusing on its effectiveness and memory efficiency.
		
		\paragraph{ZO-BOLT in practice.}
		ZO-BOLT applies ZO optimization to BOLT's fixed weighted-SFT objective under the matched-data regime, incorporating a symmetric finite-difference estimator, ZoAR query reuse~\cite{qiu2025zerothorderoptimizationsecretlysinglestep}, and R-AdaZO~\cite{shu2025refining} for improved optimization efficiency. Algorithm~\ref{alg:zobolt} outlines the method, with implementation details provided in App.~\ref{app:zobolt_experiments}. Its relationship to the forward estimator analyzed in~\eqref{eq:gradient_estimator} is discussed in App.~\ref{app:optimizer_theory}.

		\begin{algorithm}[H]
		\caption{ZO-BOLT with Symmetric Finite Differences, ZoAR, and R-AdaZO.}
		\label{alg:zobolt}
		\begin{algorithmic}[1]
		\Require \textcolor{red!75!black}{\textbf{Fixed dataset} $\mathcal D$}; initial parameters $\boldsymbol\theta_0$; perturbation scale $\mu$; minibatch size $B$; history size $N_h$; steps $T$; learning rates $\{\eta_t\}$; R-AdaZO settings
		\State Initialize ZoAR query history and R-AdaZO optimizer state
		\For{$t=1,\ldots,T$}
		\State Sample minibatch $\mathcal B_t$ of size $B$ and Gaussian direction $\mathbf u_t\sim\mathcal N(0,I_d)$
		\State $y_t^+\gets F_{\textcolor{red!75!black}{\mathcal B_t}}(\boldsymbol\theta_{t-1}+\mu\mathbf u_t)$
		\State $y_t^-\gets F_{\textcolor{red!75!black}{\mathcal B_t}}(\boldsymbol\theta_{t-1}-\mu\mathbf u_t)$ 
		\Comment{\textcolor{red!75!black}{\textbf{Same minibatch $\mathcal B_t$ for both function evaluations}}}
		\State $\mathbf g_t\gets\operatorname{ZoARUpdate}(\mathbf u_t,y_t^+,y_t^-;N_h)$
		\State $\boldsymbol\theta_t\gets\operatorname{R\!\text{-}AdaZOUpdate}(\boldsymbol\theta_{t-1},\mathbf g_t;\eta_t)$
		\EndFor
		\State \Return $\boldsymbol\theta_T$
		\end{algorithmic}
		\end{algorithm}

		\paragraph{Performance across models and tasks.}
		Table~\ref{tab:generalization} compares ZO-BOLT with the pretrained baseline and first-order BOLT on GSM8K and MATH across Qwen2.5-1.5B, Qwen2.5-7B, and DeepSeek-R1-Distill-Qwen-1.5B. All methods use LoRA ($r=8,\alpha=16$). App.~\ref{app:zobolt_experiments} provides the training details.
		
		\begin{table}[ht]
			\caption{\textbf{ZO-BOLT performance across tasks and model variants.} The table reports best-checkpoint accuracy for trained methods. Values are shown as ``Accuracy ($\Delta$)'', where $\Delta$ is the improvement over the pretrained baseline. Rec. (\%) denotes the percentage of the first-order BOLT gain recovered by ZO-BOLT. Gains are shown in red.}
			\label{tab:generalization}
			\centering
			\renewcommand{\arraystretch}{1.2}
			\setlength{\tabcolsep}{2.5pt}
			\small
			\begin{tabular}{cccccc}
				\toprule
				\textbf{Model} & \textbf{Task} & \textbf{Baseline} & \textbf{BOLT (FO)} & \textbf{ZO-BOLT (ZO)} & \textbf{Rec. (\%)} \\
				\midrule
				Qwen2.5-1.5B & GSM8K & 52.24 & 63.99~\textcolor{red!75!black}{$(+11.75)$} & 63.78~\textcolor{red!75!black}{$(+11.54)$} & 98.2 \\
				Qwen2.5-1.5B & MATH  & 29.82 & 33.60~\textcolor{red!75!black}{$(+3.78)$}  & 32.84~\textcolor{red!75!black}{$(+3.02)$}  & 79.9 \\
				\cmidrule{1-6}
				Qwen2.5-7B   & GSM8K & 75.93 & 86.56~\textcolor{red!75!black}{$(+10.63)$} & 85.73~\textcolor{red!75!black}{$(+9.80)$}  & 92.2 \\
				Qwen2.5-7B   & MATH  & 50.30 & 61.42~\textcolor{red!75!black}{$(+11.12)$} & 58.00~\textcolor{red!75!black}{$(+7.70)$}  & 69.2 \\
				\cmidrule{1-6}
				DeepSeek-R1-Distill-Qwen-1.5B & GSM8K & 60.97 & 70.77~\textcolor{red!75!black}{$(+9.80)$}  & 66.89~\textcolor{red!75!black}{$(+5.92)$}  & 60.4 \\
				DeepSeek-R1-Distill-Qwen-1.5B & MATH  & 52.28 & 54.66~\textcolor{red!75!black}{$(+2.38)$}  & 52.78~\textcolor{red!75!black}{$(+0.50)$}  & 21.0 \\
				\bottomrule
			\end{tabular}
		\end{table}
		
		ZO-BOLT consistently improves over the pretrained model across all six settings, recovering a substantial portion of the corresponding first-order BOLT gains on GSM8K, with more limited recovery on the more challenging MATH benchmark. These results demonstrate the effectiveness of ZO-BOLT across different models and mathematical reasoning benchmarks.
		
		\paragraph{Memory efficiency.}
		We assess the memory efficiency of ZO-BOLT by comparing peak GPU memory and inference overhead against BOLT, GRPO, and ZO-GRPO at the 1.5B and 7B scales under consistent experimental settings (Table~\ref{tab:memory_combined}). Experimental details are provided in App.~\ref{app:memory_experiments}.
		
		\begin{table}[ht]
			\caption{\textbf{Peak GPU memory usage} on NVIDIA L40 GPUs under LoRA ($r=8$, $\alpha=16$, bs$=2$, max\_len$=1024$, bf16). Best and second-best values among training methods are bolded and underlined.}
			\label{tab:memory_combined}
			\centering
			\begin{tabular}{lcccc}
				\toprule
				\multicolumn{1}{c}{\textbf{Method}} & \multicolumn{2}{c}{\textbf{Qwen2.5-1.5B}} & \multicolumn{2}{c}{\textbf{Qwen2.5-7B}} \\
				\cmidrule(r){2-3} \cmidrule(r){4-5}
				& \textbf{Peak (MB)} & \textbf{$\Delta$ over Inference} & \textbf{Peak (MB)} & \textbf{$\Delta$ over Inference} \\
				\midrule
				Inference only & 3{,}281 & -- & 14{,}900 & -- \\
				BOLT (FO)       & 10{,}363 & $+7{,}082$ & 28{,}436 & $+13{,}536$ \\
				GRPO (FO)      & 12{,}012 & $+8{,}731$ & 29{,}379 & $+14{,}479$ \\
				\textbf{ZO-BOLT}         & \textbf{4{,}201}  & $\boldsymbol{+920}$     & \textbf{16{,}014} & $\boldsymbol{+1{,}114}$  \\
				ZO-GRPO (ZO)   & \underline{5{,}245}  & $\underline{+1{,}964}$ & \underline{16{,}598} & $\underline{+1{,}698}$  \\
				\bottomrule
			\end{tabular}
		\end{table}
		
		ZO-BOLT reduces peak GPU memory by $59.5\%$ and $43.7\%$ compared with first-order BOLT at the 1.5B and 7B scales, respectively, while maintaining a near-inference memory footprint. These results confirm that ZO-BOLT retains the memory efficiency of standard ZO optimization alongside its demonstrated performance gains.
		
		\section{Conclusions and Limitations}
		\label{sec:conclusion}
		
		This work establishes that ZO gradient reliability is structurally determined by the data regime: on-policy RL under the mismatched-data regime incurs a $1/\mu^2$ oracle-noise amplification that sets an irreducible noise floor, while BOLT under the matched-data regime eliminates this amplification. Controlled variance measurements and matched-versus-mismatched end-to-end comparisons empirically validate this separation. Guided by these findings, ZO-BOLT provides a practical realization of matched-data ZO optimization toward the corresponding RLVR-aligned target, achieving consistent improvements across tasks and model variants while retaining a near-inference memory footprint.
		
		\paragraph{Limitations and Future Work.}
		Our experiments are limited to mathematical reasoning tasks and models up to 7B parameters, and the performance gap between zeroth- and first-order BOLT widens on harder tasks. Future work will pursue broader task coverage, larger model scales, and improved zeroth-order optimization strategies to narrow these gaps.
		
		\section*{Acknowledgments}
		This work is supported by National Natural Science Foundation of China (U22B2017), International Cooperation Foundation of Hubei Province, China (2024EHA032), and the Youth S\&T Talent Support Programme of Guangdong Provincial Association for Science and Technology (Grant No. SKXRC2025466).
		
		\medskip
		
		\bibliographystyle{unsrtnat}
		\bibliography{references}
		
		\newpage
		\appendix
		
		\section{Related Work}
		\label{app:related_work}
		
		\paragraph{Zeroth-order optimization for LLM fine-tuning.}
		
		Zeroth-order (ZO) optimization has recently emerged as a memory-efficient paradigm for fine-tuning LLMs, owing to its backpropagation-free nature. MeZO~\cite{malladi2023fine} first established the feasibility of this paradigm by demonstrating that billion-parameter LLMs can be fine-tuned via forward-only updates with an inference-level memory footprint in supervised fine-tuning (SFT) setting. Subsequent work has focused on improving convergence and reducing gradient variance through low-dimensional perturbations~\cite{yu2025zeroth,chen2026ar1zo}, sparse updates~\cite{liu2025sparseMeZO}, variance reduction~\cite{gautam2024variance}, query reuse~\cite{qiu2025zerothorderoptimizationsecretlysinglestep,feng2026zoaq}, and adaptive strategies~\cite{shu2025refining}, with recent extensions to reinforcement learning from human feedback (RLHF)~\cite{zhang2025zeroth} and LLM preference optimization~\cite{galatolo2025visualising}. These results establish the feasibility of memory-efficient ZO post-training, but primarily study fixed-data or parameter-independent sampling settings. To the best of our knowledge, the applicability of ZO optimization to RLVR remains largely unexplored. This gap motivates examining how policy-dependent data generation affects the paired function evaluations used by the ZO gradient estimator.
		
		\paragraph{Mismatched-data regime: On-policy RL.}
		
		Policy-gradient optimization is grounded in the Policy Gradient Theorem~\cite{sutton1999policy}, with REINFORCE~\cite{williams1992simple} as a foundational algorithm. Early policy gradient methods suffer from high variance and limited sample efficiency~\cite{greensmith2004variance}, motivating methods such as TRPO~\cite{schulman2015trust}, PPO~\cite{schulman2017proximal}, and GRPO~\cite{shao2024deepseekmath}, which improve optimization stability and sample efficiency through constrained or clipped policy updates and improved advantage estimation~\cite{murphy2024reinforcement, robert2004monte}. However, applying ZO optimization directly to an on-policy objective with independently sampled rollouts places the paired function evaluations in the \emph{mismatched-data regime}, introducing stochastic differences that can amplify ZO gradient estimation variance. Despite extensive research on policy-gradient variance reduction, the impact of this regime on the oracle-noise variance of ZO gradient estimators remains insufficiently characterized.
		
		\paragraph{Matched-data regime: Fixed-data optimization.}
		To reduce the computational cost of online sampling, alternative approaches optimize objectives constructed from offline data~\cite{levine2020offline,mu2026drift}. In the Reinforcement Learning from Human Feedback (RLHF) setting, DPO~\cite{rafailov2023direct} enables direct preference optimization on fixed preference pairs without explicit reward-model training or online sampling, with iterative extensions~\cite{pang2024iterative, chen2024self} further expanding data coverage across training rounds. In the Reinforcement Learning with Verifiable Rewards (RLVR) setting, BOLT~\cite{shu2026referencesampledboltzmannprojectionklregularized} identifies the Boltzmann target of fixed-reference KL-regularized RLVR and constructs a reference-sampled, prompt-normalized weighted-SFT objective whose induced policy matches this target at the population level. This construction enables more efficient multi-epoch training on fixed-data, significantly reducing memory overhead and accelerating training time. While existing studies primarily focus on first-order optimization, the role of matched data in mitigating oracle-noise amplification under ZO optimization remains underexplored. In this work, we characterize this advantage of the matched-data regime through BOLT and establish corresponding theoretical guarantees.

		\section{Limitations of Variance Reduction for ZO Optimization in On-Policy RL}
		\subsection{Limitations of Sample Reuse for Variance Reduction}
		\label{app:rl:reuse}
		
		A natural question is whether reusing the same trajectories across evaluations can eliminate the $1/\mu^2$ amplification. To formalize this idea, we introduce the \emph{frozen-distribution surrogate} $\bar J_\theta(\theta')$, which evaluates the loss at new parameters $\theta'$ using trajectories drawn from the original parameter distribution $P_\theta$. Crucially, the following Proposition~\ref{prop:frozen_surrogate} (proof in App.~\ref{app:proof_frozen_surrogate}) presents an exact decomposition of the on-policy objective, showing that any variance reduction from trajectory reuse comes at the cost of introducing a distribution-shift term.
		
		\begin{proposition}[Frozen-sample reuse induces a different objective]
			\label{prop:frozen_surrogate}
			Define the frozen-distribution surrogate
			\begin{equation}
				\bar J_\theta(\theta')
				\;:=\;
				\mathbb{E}_{\tau \sim P_\theta}[\ell(\theta';\tau)].
			\end{equation}
			
			Then for every $\theta, u \in \mathbb{R}^d$ and $\mu > 0$, the following identity holds:
			\begin{equation}
				\label{eq:frozen_decomposition}
				J(\theta+\mu u)-J(\theta)
				=
				\underbrace{\big(\bar J_\theta(\theta+\mu u)-\bar J_\theta(\theta)\big)}_{\text{same-distribution term}}
				+
				\underbrace{\big(J(\theta+\mu u)-\bar J_\theta(\theta+\mu u)\big)}_{\text{distribution-shift term}}.
			\end{equation}
		\end{proposition}
		
		\paragraph{Remark.}
		This identity reveals an inherent tension between sample reuse and optimization fidelity. Sample reuse reduces finite-difference noise, but shifts the optimization target from the true on-policy objective $J$ to the frozen-distribution surrogate $\bar{J}_\theta$. The distribution-shift term $J(\theta+\mu u)-\bar{J}_\theta(\theta+\mu u)$ vanishes if and only if $P_{\theta+\mu u} = P_\theta$. In other words, variance reduction through sample reuse inherently comes at the cost of changing the optimization target.
		
		\subsection{Importance Weighting and Exponential Blow-up}
		\label{app:rl:importance}
		To quantify the cost of correcting the distribution-shift bias identified above, we next examine \emph{importance sampling}, the standard technique for reweighting samples from one distribution to target another. Applied to the frozen-distribution surrogate, this reweights trajectories to target the true on-policy RL objective:
		
		\begin{equation}
			\label{eq:importance}
			J(\theta')
			=
			\mathbb{E}_{\tau \sim P_\theta}
			\left[
			\frac{P_{\theta'}(\tau)}{P_\theta(\tau)}\,\ell(\theta';\tau)
			\right],
			\quad
			\frac{P_{\theta'}(\tau)}{P_\theta(\tau)}
			=
			\prod_{t=1}^T
			\frac{\pi_{\theta'}(y_t|y_{<t},x)}{\pi_\theta(y_t|y_{<t},x)}.
		\end{equation}
		While this correction is unbiased, it comes with a hidden cost: the second moment of the importance weights can grow exponentially with the sequence length $T$. 
		The following Corollary~\ref{cor:importance_weight_second_moment} (proof in App.~\ref{app:proof_importance_weight}) formalizes this structural limitation.
		
		\begin{corollary}[Importance-weight second moment under uniform overlap]
			\label{cor:importance_weight_second_moment}
			
			Fix $\theta, \theta'$. Let
			\begin{equation}
				w_{\theta \to \theta'}(\tau) := \frac{P_{\theta'}(\tau)}{P_\theta(\tau)}, \quad \tau = (y_1, \dots, y_T).
			\end{equation}
			
			Assume $P_{\theta'} \ll P_\theta$, and suppose there exists $\rho \ge 1$ such that for every history $h_t = (x, y_{<t})$,
			\begin{equation}
				\sum_{y_t} \frac{\pi_{\theta'}(y_t \mid h_t)^2}{\pi_\theta(y_t \mid h_t)} \le \rho.
			\end{equation}
			
			Then
			\begin{equation}
				\mathbb{E}_{\tau \sim P_\theta}\!\left[w_{\theta \to \theta'}(\tau)^2\right] \le \rho^T.
			\end{equation}
			
		\end{corollary}
		
		\paragraph{Remark.}
		Corollaries~\ref{cor:independent_noise} and~\ref{cor:importance_weight_second_moment} together reveal a fundamental trade-off: exact sample reuse preserves low variance only for the frozen surrogate $\bar J_\theta$. Correcting back to the true objective requires importance weighting, but its variance can increase exponentially with sequence length. This establishes a structural tension between unbiasedness and variance control in ZO optimization for on-policy RL.
		
		\subsection{PPO/GRPO-Style Frozen-Distribution Surrogate Methods}
		\label{app:rl:ppo}
		Methods such as PPO~\cite{schulman2017proximal} and GRPO~\cite{shao2024deepseekmath} implement a frozen-distribution surrogate strategy consistent with Prop.~\ref{prop:frozen_surrogate}. These methods first sample trajectories from a frozen behavior policy $\pi_{\mathrm{old}}$ and then optimize a surrogate objective on that fixed batch. As a result, the two ZO evaluations can reuse identical samples, producing oracle-noise behavior closer to a matched-data regime. However, this improvement comes at the cost highlighted in Prop.~\ref{prop:frozen_surrogate}: the optimization target is no longer the true on-policy objective \(
		J(\theta) = \mathbb{E}_{\tau \sim P_\theta}[\ell(\theta;\tau)]
		\), but a frozen-distribution surrogate defined relative to $P_{\pi_{\mathrm{old}}}$, often with additional modifications such as clipping or local trust-region constraints. 
		
		Moreover, if the surrogate employs token-level or clipped importance ratios that aggregate additively rather than via the exact trajectory-level product, the exponential growth in the second moment identified in Cor.~\ref{cor:importance_weight_second_moment} no longer holds strictly. This occurs because the method intentionally departs from exact unbiased correction for $J(\theta)$.
		
		In summary, PPO and GRPO achieve variance reduction by transitioning to a frozen-distribution surrogate, but this is achieved by changing the optimization target, \emph{not by resolving the fundamental structural mismatch present in direct on-policy RL}.
		
		\section{Proofs}
		\label{sec:appendix_proofs}
		
		\subsection{Proof of Lemma~\ref{lem:noise_identity}}
		\label{app:proof_noise_identity}
		\begin{proof}
			By the decomposition in Section~\ref{sec:framework},
			\begin{equation}
				\label{eq:proof_decomp}
				\hat g - g_F
				=
				\frac{e(\theta+\mu u,\xi_+) - e(\theta,\xi_-)}{\mu}\, u.
			\end{equation}
			
			Taking squared norms gives
			\begin{equation}
				\label{eq:proof_norm}
				\|\hat g - g_F\|^2
				=
				\frac{\bigl(e(\theta+\mu u,\xi_+) - e(\theta,\xi_-)\bigr)^2}{\mu^2}\,\|u\|^2.
			\end{equation}
			
			Taking conditional expectation with respect to $\theta$ yields~\eqref{eq:noise_identity}.
			
			For the upper bound, condition on $u$ and apply the bound on the conditional second moment:
			\begin{align}
				\mathbb{E}\!\left[\|\hat g - g_F\|^2 \,\middle|\, \theta\right]
				&=
				\frac{1}{\mu^2}
				\mathbb{E}_u\!\left[
				\|u\|^2\,
				\mathbb{E}\!\left[
				\bigl(e(\theta+\mu u,\xi_+) - e(\theta,\xi_-)\bigr)^2
				\,\middle|\, \theta, u
				\right]
				\right] \\
				&\le
				\frac{\nu^2}{\mu^2}\,
				\mathbb{E}[\|u\|^2]
				=
				\frac{d\,\nu^2}{\mu^2},
			\end{align}
			which proves the claimed upper bound.
		\end{proof}
		
		\subsection{Proof of Corollary~\ref{cor:independent_noise}}
		\label{app:proof_independent_noise}
		
		\begin{proof}
			By conditional independence,
			\begin{equation}
				\operatorname{Var}\bigl(e(\theta+\mu u,\xi_+) - e(\theta,\xi_-)\mid \theta,u\bigr) \le 2\sigma^2.
			\end{equation}
			
			Since
			\begin{equation}
				\mathbb{E}\!\left[e(\theta+\mu u,\xi_+) - e(\theta,\xi_-)\mid \theta, u\right] = 0
			\end{equation}
			under Assumption~\ref{ass:oracle}, the second moment equals the variance.  
			
			Applying Lemma~\ref{lem:noise_identity} with $\nu^2 = 2\sigma^2$ gives
			\begin{equation}
				\mathbb{E}\!\left[\|\hat g - g_F\|^2 \,\middle|\, \theta\right]
				\le
				\frac{2\,d\,\sigma^2}{\mu^2},
			\end{equation}
			which proves~\eqref{eq:independent_noise}.
		\end{proof}
		
		\subsection{Proof of Proposition~\ref{prop:online_rl_noise}}
		\label{app:proof_online_rl_noise}
		
		\begin{proof}
			This is a direct specialization of Corollary~\ref{cor:independent_noise} with 
			\(
			F = J
			\) 
			and 
			\(
			\sigma^2 = \sigma_{\mathrm{roll}}^2
			\).  
			
			The on-policy oracle \(\tilde J(\theta, \xi)\) satisfies Assumption~\ref{ass:oracle} via Assumption~\ref{ass:online_oracle} (unbiasedness).  
			
			The conditional independence of rollouts at \(\theta\) and \(\theta + \mu u\) holds because on-policy sampling generates fresh trajectories independently at each parameter value.  
			
			Applying Corollary~\ref{cor:independent_noise} then gives
			\begin{equation}
				\mathbb{E}\!\left[\|\hat g_{\mathrm{on\_policy}} - g_J\|^2 \,\middle|\, \theta\right]
				\le
				\frac{2\,d\,\sigma_{\mathrm{roll}}^2}{\mu^2},
			\end{equation}
			which proves~\eqref{eq:online_rl_bound}.
		\end{proof}
		
		\subsection{Proof of Proposition~\ref{prop:frozen_surrogate}}
		\label{app:proof_frozen_surrogate}
		
		\begin{proof}
			By definition of $J(\theta)$ and $\bar J_\theta(\theta)$,
			\begin{equation}
				\bar J_\theta(\theta)
				=
				\mathbb{E}_{\tau \sim P_\theta}[\ell(\theta;\tau)]
				=
				J(\theta).
			\end{equation}
			
			Now add and subtract $\bar J_\theta(\theta+\mu u)$:
			\begin{align}
				J(\theta+\mu u)-J(\theta)
				&=
				\big(J(\theta+\mu u)-\bar J_\theta(\theta+\mu u)\big) \\
				&\quad+
				\big(\bar J_\theta(\theta+\mu u)-\bar J_\theta(\theta)\big) \\
				&\quad+
				\big(\bar J_\theta(\theta)-J(\theta)\big).
			\end{align}
			
			The last term vanishes since $\bar J_\theta(\theta)=J(\theta)$, yielding
			\begin{equation}
				J(\theta+\mu u)-J(\theta)
				=
				\big(\bar J_\theta(\theta+\mu u)-\bar J_\theta(\theta)\big)
				+
				\big(J(\theta+\mu u)-\bar J_\theta(\theta+\mu u)\big),
			\end{equation}
			which proves~\eqref{eq:frozen_decomposition}.
		\end{proof}
		
		\subsection{Proof of Corollary~\ref{cor:importance_weight_second_moment}}
		\label{app:proof_importance_weight}
		
		\begin{proof}
			Write 
			\begin{equation}
				r_t := \frac{\pi_{\theta'}(y_t \mid h_t)}{\pi_\theta(y_t \mid h_t)},
			\end{equation}
			so that 
			\begin{equation}
				w_{\theta \to \theta'}(\tau)^2 = \prod_{t=1}^T r_t^2.
			\end{equation}
			
			By the tower property,
			\begin{equation}
				\mathbb{E}[w_{\theta \to \theta'}(\tau)^2]
				=
				\mathbb{E}\!\left[\prod_{t=1}^{T-1} r_t^2 \cdot \mathbb{E}[r_T^2 \mid h_T]\right]
				\le
				\rho \cdot \mathbb{E}\!\left[\prod_{t=1}^{T-1} r_t^2\right].
			\end{equation}
			
			Iterating this argument $T$ times yields
			\begin{equation}
				\mathbb{E}_{\tau \sim P_\theta}[w_{\theta \to \theta'}(\tau)^2] \le \rho^T,
			\end{equation}
			which proves Corollary~\ref{cor:importance_weight_second_moment}.
		\end{proof}

		\subsection{Proof of Proposition~\ref{prop:bolt_variance}}
		\label{app:proof_bolt}
		
		\begin{proof}
			We provide the full derivation.
			
			\paragraph{Step 1: Decomposition.}
			Define
			\begin{equation}
				\label{eq:app_delta_def}
				\Delta_{\mathcal{B}}(\theta,u)
				:=
				\big(F_{\mathcal{B}}(\theta+\mu u) - F_{\mathcal{B}}(\theta)\big)
				-
				\big(F_{\mathrm{BOLT}}(\theta+\mu u) - F_{\mathrm{BOLT}}(\theta)\big).
			\end{equation}
			Then
			\begin{equation}
				\label{eq:app_decomp}
				\hat g_{\mathrm{BOLT}} - g_{F_{\mathrm{BOLT}}}
				=
				\frac{\Delta_{\mathcal{B}}(\theta,u)}{\mu}\,u.
			\end{equation}
			
			\paragraph{Step 2: Fundamental theorem of calculus.}
			Let
			\(
			\delta_{\mathcal{B}}(\theta)
			:=
			\nabla F_{\mathcal{B}}(\theta) - \nabla F_{\mathrm{BOLT}}(\theta).
			\)
			By the fundamental theorem of calculus,
			\begin{equation}
				\label{eq:app_ftc}
				\Delta_{\mathcal{B}}(\theta,u)
				=
				\mu\langle \delta_{\mathcal{B}}(\theta),\,u\rangle
				+
				R(\theta,u,\mathcal{B}),
			\end{equation}
			where
			\begin{equation}
				\label{eq:app_remainder}
				R(\theta,u,\mathcal{B})
				=
				\mu\int_0^1
				\langle
				\delta_{\mathcal{B}}(\theta+t\mu u) - \delta_{\mathcal{B}}(\theta),
				\,u
				\rangle
				\,dt.
			\end{equation}
			
			\paragraph{Step 3: Remainder bound.}
			Since both $F_{\mathcal{B}}$ and $F_{\mathrm{BOLT}}$ are $L$-smooth (Assumption~\ref{ass:smoothness}),
			$\delta_{\mathcal{B}}$ is $2L$-Lipschitz. Hence
			\begin{equation}
				\label{eq:app_R_bound}
				|R(\theta,u,\mathcal{B})|
				\le
				\mu\int_0^1 2Lt\mu\|u\|^2\,dt
				=
				L\mu^2\|u\|^2.
			\end{equation}
			
			\paragraph{Step 4: Minkowski decomposition.}
			Substituting~\eqref{eq:app_ftc} into~\eqref{eq:app_decomp} gives
			\begin{equation}
				\label{eq:app_main_decomp}
				\hat g_{\mathrm{BOLT}} - g_{F_{\mathrm{BOLT}}}
				=
				\langle \delta_{\mathcal{B}}(\theta),\,u\rangle\,u
				+
				\frac{R(\theta,u,\mathcal{B})}{\mu}\,u.
			\end{equation}
			Applying Minkowski’s inequality in $L_2$,
			\begin{align}
				\label{eq:app_minkowski}
				\Big(\mathbb{E}\|\hat g_{\mathrm{BOLT}} - g_{F_{\mathrm{BOLT}}}\|^2\Big)^{1/2}
				&\le
				\Big(
				\mathbb{E}\big[\langle \delta_{\mathcal{B}}(\theta),u\rangle^2\|u\|^2\big]
				\Big)^{1/2}
				\nonumber\\
				&\quad+
				\Big(
				\mathbb{E}\big[R^2\|u\|^2/\mu^2\big]
				\Big)^{1/2}.
			\end{align}
			
			\paragraph{Step 5: First term.}
			Condition on $\mathcal{B}$ and use the Gaussian moment identity
			\(
			\mathbb{E}_u[\langle v, u\rangle^2 \|u\|^2] = (d+2)\|v\|^2
			\)
			for $u \sim \mathcal{N}(0,I_d)$. Taking expectation over $\mathcal{B}$ and applying Assumption~\ref{ass:minibatch_variance},
			\begin{equation}
				\label{eq:app_first_term}
				\mathbb{E}\!\left[
				\langle \delta_{\mathcal{B}}(\theta),u\rangle^2\|u\|^2
				\right]
				=
				(d+2)\,\mathbb{E}_{\mathcal{B}}\!\left[\|\delta_{\mathcal{B}}(\theta)\|^2\right]
				\le
				\frac{(d+2)\,\sigma_{\mathrm{mb}}^2}{B}.
			\end{equation}
			Hence
			\begin{equation}
				\label{eq:app_first_term_sqrt}
				\Big(
				\mathbb{E}[\langle \delta_{\mathcal{B}},u\rangle^2\|u\|^2]
				\Big)^{1/2}
				\le
				\sqrt{\frac{(d+2)\sigma_{\mathrm{mb}}^2}{B}}.
			\end{equation}
			
			\paragraph{Step 6: Second term.}
			Using~\eqref{eq:app_R_bound} and $\mathbb{E}[\|u\|^6] = d(d+2)(d+4)$,
			\begin{equation}
				\label{eq:app_second_term}
				\mathbb{E}\!\left[\frac{R^2}{\mu^2}\|u\|^2\right]
				\le
				L^2\mu^2\,\mathbb{E}[\|u\|^6]
				=
				L^2\mu^2\,d(d+2)(d+4).
			\end{equation}
			Thus
			\begin{equation}
				\label{eq:app_second_term_sqrt}
				\Big(
				\mathbb{E}[R^2\|u\|^2/\mu^2]
				\Big)^{1/2}
				\le
				L\mu\sqrt{d(d+2)(d+4)}.
			\end{equation}
			
			\paragraph{Step 7: Combine.}
			Substituting~\eqref{eq:app_first_term_sqrt} and~\eqref{eq:app_second_term_sqrt} into~\eqref{eq:app_minkowski} yields
			\begin{equation}
				\label{eq:app_final_bound}
				\Big(\mathbb{E}\|\hat g_{\mathrm{BOLT}} - g_{F_{\mathrm{BOLT}}}\|^2\Big)^{1/2}
				\le
				\sqrt{\frac{(d+2)\sigma_{\mathrm{mb}}^2}{B}}
				+
				L\mu\sqrt{d(d+2)(d+4)}.
			\end{equation}
			
			Squaring gives the stated bound.
		\end{proof}

		\subsection{Proof of Lemma~\ref{lem:smoothing}}
		\label{app:proof_smoothing}
		\begin{proof}
			The first identity follows from the Gaussian smoothing formula
			\begin{equation}
				\nabla F_\mu(\theta) = \frac{1}{\mu}\mathbb{E}\bigl[F(\theta+\mu u)\,u\bigr],
			\end{equation}
			together with $\mathbb{E}[u]=0$.
			
			For the second identity, we write
			\begin{equation}
				\nabla F_\mu(\theta) - \nabla F(\theta)
				= \mathbb{E}\bigl[\nabla F(\theta+\mu u) - \nabla F(\theta)\bigr].
			\end{equation}
			Applying $L_F$-smoothness and Jensen's inequality yields the stated bound.
		\end{proof}
		
		\subsection{Proof of Lemma~\ref{lem:zo_second_moment}}
		\label{app:proof_zo_second_moment}
		\begin{proof}
			By smoothness,
			\begin{equation}
				F(\theta+\mu u) - F(\theta)
				= \mu \langle \nabla F(\theta), u\rangle + r(\theta,u),
			\end{equation}
			where
			\begin{equation}
				|r(\theta,u)| \le \frac{L_F \mu^2}{2}\|u\|^2.
			\end{equation}
			Hence,
			\begin{equation}
				g_F(\theta;u)
				= \langle \nabla F(\theta), u\rangle\,u + \frac{r(\theta,u)}{\mu}\,u.
			\end{equation}
			
			Using $(a+b)^2 \le 2a^2 + 2b^2$, we obtain
			\begin{equation}
				\|g_F(\theta;u)\|^2
				\le 2\langle \nabla F(\theta), u\rangle^2 \|u\|^2
				+ \frac{L_F^2 \mu^2}{2}\|u\|^6.
			\end{equation}
			
			Taking expectations and using
			\begin{equation}
				\mathbb{E}\bigl[\langle v,u\rangle^2\|u\|^2\bigr] = (d+2)\|v\|^2,
				\quad
				\mathbb{E}\bigl[\|u\|^6\bigr] = d(d+2)(d+4),
			\end{equation}
			we get
			\begin{equation}
				\mathbb{E}\|g_F(\theta;u)\|^2
				\le 2(d+2)\|\nabla F(\theta)\|^2
				+ \frac{L_F^2 \mu^2}{2}d(d+2)(d+4).
			\end{equation}
			
			By Lemma~\ref{lem:smoothing},
			\begin{equation}
				\|\nabla F(\theta)\|^2
				\le 2\|\nabla F_\mu(\theta)\|^2 + 2L_F^2 \mu^2 d.
			\end{equation}
			Substituting gives
			\begin{equation}
				\mathbb{E}\|g_F(\theta;u)\|^2
				\le 4(d+2)\|\nabla F_\mu(\theta)\|^2
				+ 4(d+2)L_F^2 \mu^2 d
				+ \frac{L_F^2 \mu^2}{2}d(d+2)(d+4).
			\end{equation}
			
			Collecting the $L_F^2 \mu^2 d(d+2)$ terms yields the stated bound.
		\end{proof}
		
		\subsection{Proof of Theorem~\ref{thm:zooconvergence}}
		\label{app:proof_zooconvergence}
		\begin{proof}
			The proof uses the smoothed-objective descent framework with an explicit decomposition of the estimator's second moment via Lemma~\ref{lem:smoothing} and Lemma~\ref{lem:zo_second_moment}.
			
			\textbf{Step 1: Unbiasedness.}
			By Lemma~\ref{lem:smoothing} and the conditional unbiasedness assumption,
			\begin{equation}
				\mathbb{E}[\hat g_t \mid \theta_t]
				= \mathbb{E}[g_F(\theta_t;u_t) \mid \theta_t]
				= \nabla F_\mu(\theta_t).
			\end{equation}
			
			\textbf{Step 2: One-step progress.}
			Since $F$ is $L_F$-smooth, its Gaussian smoothing $F_\mu$ is also $L_F$-smooth (Nesterov \& Spokoiny, 2017, Lemma 1). For the update $\theta_{t+1} = \theta_t - \eta \hat g_t$,
			\begin{equation}
				F_\mu(\theta_{t+1})
				\le
				F_\mu(\theta_t)
				- \eta \langle \nabla F_\mu(\theta_t), \hat g_t \rangle
				+ \frac{L_F \eta^2}{2}\|\hat g_t\|^2.
			\end{equation}
			
			Taking expectation conditioned on $\theta_t$ and using Step 1,
			\begin{equation}
				\mathbb{E}[F_\mu(\theta_{t+1}) \mid \theta_t]
				\le
				F_\mu(\theta_t)
				- \eta \|\nabla F_\mu(\theta_t)\|^2
				+ \frac{L_F \eta^2}{2}\mathbb{E}[\|\hat g_t\|^2 \mid \theta_t].
			\end{equation}
			
			\textbf{Step 3: Second moment decomposition.}
			Write $\hat g_t = g_F(\theta_t;u_t) + \zeta_t$, where $\zeta_t := \hat g_t - g_F(\theta_t;u_t)$. Since $\mathbb{E}[\zeta_t \mid \theta_t, u_t] = 0$, the cross term vanishes, giving
			\begin{equation}
				\mathbb{E}[\|\hat g_t\|^2 \mid \theta_t]
				=
				\mathbb{E}[\|g_F(\theta_t;u_t)\|^2 \mid \theta_t]
				+
				\mathbb{E}[\|\zeta_t\|^2 \mid \theta_t].
			\end{equation}
			
			\textbf{Step 4: Apply Lemma~\ref{lem:zo_second_moment}.}
			By Lemma~\ref{lem:zo_second_moment},
			\begin{equation}
				\mathbb{E}[\|g_F(\theta_t;u_t)\|^2 \mid \theta_t]
				\le
				4(d+2)\|\nabla F_\mu(\theta_t)\|^2
				+
				\frac{L_F^2 \mu^2\, d(d+2)(d+12)}{2}.
			\end{equation}
			
			Combining with $\mathbb{E}[\|\zeta_t\|^2 \mid \theta_t] \le V$,
			\begin{equation}
				\mathbb{E}[\|\hat g_t\|^2 \mid \theta_t]
				\le
				4(d+2)\|\nabla F_\mu(\theta_t)\|^2
				+
				V
				+
				\frac{L_F^2 \mu^2\, d(d+2)(d+12)}{2}.
			\end{equation}
			
			\textbf{Step 5: Substitute into descent.}
			Plugging Step 4 into Step 2,
			\begin{equation}
				\mathbb{E}[F_\mu(\theta_{t+1}) \mid \theta_t]
				\le
				F_\mu(\theta_t)
				- \eta \bigl(1 - 2L_F\eta(d+2)\bigr)\|\nabla F_\mu(\theta_t)\|^2
				+ \frac{L_F\eta^2}{2}
				\left(
				V + \frac{L_F^2 \mu^2\, d(d+2)(d+12)}{2}
				\right).
			\end{equation}
			
			\textbf{Step 6: Step size condition.}
			When $\eta \le \frac{1}{4L_F(d+2)}$, we have
			\begin{equation}
				1 - 2L_F\eta(d+2) \ge \frac{1}{2}.
			\end{equation}
			
			Thus,
			\begin{equation}
				\mathbb{E}[F_\mu(\theta_{t+1})]
				\le
				\mathbb{E}[F_\mu(\theta_t)]
				- \frac{\eta}{2}\mathbb{E}\|\nabla F_\mu(\theta_t)\|^2
				+ \frac{L_F\eta^2}{2}
				\left(
				V + \frac{L_F^2 \mu^2\, d(d+2)(d+12)}{2}
				\right).
			\end{equation}
			
			\textbf{Step 7: Telescope.}
			Summing over $t=0,\dots,T-1$ and rearranging,
			\begin{equation}
				\frac{1}{T}\sum_{t=0}^{T-1}\mathbb{E}\|\nabla F_\mu(\theta_t)\|^2
				\le
				\frac{2\Delta_{\mu,0}}{\eta T}
				+
				L_F\eta
				\left(
				V + \frac{L_F^2 \mu^2\, d(d+2)(d+12)}{2}
				\right).
			\end{equation}
			
			The specialized bounds in parts (i) and (ii) of Theorem~\ref{thm:zooconvergence} follow by substituting the respective values of $V$ from Propositions~\ref{prop:online_rl_noise} and~\ref{prop:bolt_variance}. Setting $\eta = \frac{1}{4L_F(d+2)}$ gives the maximal admissible step size.
		\end{proof}

		\section{Experimental Details and Additional Results}
		\label{app:experiments}
		\label{app:exp_setup}
		
		\subsection{Common Experimental Setup}
		\label{app:common_setup}
		
		\textbf{Model and parameter-efficient fine-tuning.}
		We employ Qwen2.5-1.5B~\cite{qwen2025qwen25technicalreport} as our primary backbone, utilizing LoRA~\cite{hu2022lora} ($r=8, \alpha=16$) targeting the query ($q_{\text{proj}}$) and key ($k_{\text{proj}}$) projections. This setup results in approximately 1M trainable parameters. We additionally evaluate Qwen2.5-7B and DeepSeek-R1-Distill-Qwen-1.5B~\cite{guo2025deepseek}.

		\textbf{Datasets.}
		Our experiments focus primarily on \textbf{GSM8K}~\cite{cobbe2021gsm8k}, a grade-school math benchmark with 7,473 training and 1,319 test problems. We also evaluate on \textbf{MATH}~\cite{hendrycks2021math}, a more challenging competition-level dataset with 5,000 test problems, to assess performance on more difficult problems.
		
		\textbf{Evaluation.}
		Evaluation employs greedy decoding (temperature $0$) with a chain-of-thought prompt template. Accuracy is measured using a mathematical-equivalence grader on the final answer in \verb|\boxed{}| format.
		
		\subsection{Validation of the Matched- and Mismatched-Data Regimes}
		\label{app:data_matching}
		
		\subsubsection{Perturbation-Scale Sweep and Oracle-Noise Variance}
		\label{app:variance_experiments}
		
		\paragraph{Shared configuration.}
		We estimate the oracle-noise variance for on-policy RL under the mismatched-data regime and BOLT under the matched-data regime, as characterized by Prop.~\ref{prop:online_rl_noise} and Prop.~\ref{prop:bolt_variance}. Both experiments use Qwen2.5-1.5B with LoRA ($r=8$, $\alpha=16$) on the query and key projections, resulting in approximately 1.09M trainable parameters. Both experiments use FP32 at the initial LoRA parameter point $\theta^\star$, without training or parameter updates. They use $K=10$ Gaussian directions, $R=32$ repetitions, and five scales
		$\mu\in\{10^{-4},3\times10^{-4},10^{-3},3\times10^{-3},10^{-2}\}$.
		
		\paragraph{ZO evaluation on the fixed BOLT objective.}
		We construct a fixed empirical objective from 2,048 weighted GSM8K prompt--response pairs. For direction $u_k$, its exact finite-population reference is
		\begin{equation}
			g_{F,k}(\mu)
			=
			\frac{F_{\mathcal D}(\theta^\star+\mu u_k)-F_{\mathcal D}(\theta^\star)}{\mu}u_k.
		\end{equation}
		For each repetition, we draw two size-16 minibatches $\mathcal B_{k,r}^{(1)}$ and $\mathcal B_{k,r}^{(2)}$. Matched BOLT uses $\mathcal B_{k,r}^{(1)}$ for both function evaluations, whereas the independent-minibatch variant, termed mismatched BOLT, uses $\mathcal B_{k,r}^{(1)}$ and $\mathcal B_{k,r}^{(2)}$ for the two evaluations, respectively. The corresponding controlled estimators are
		\begin{align}
			\hat g^{\mathrm{match}}_{k,r}(\mu)
			&=
			\frac{F_{\mathcal B_{k,r}^{(1)}}(\theta^\star+\mu u_k)
				-F_{\mathcal B_{k,r}^{(1)}}(\theta^\star)}{\mu}u_k,\\
			\hat g^{\mathrm{mismatch}}_{k,r}(\mu)
			&=
			\frac{F_{\mathcal B_{k,r}^{(1)}}(\theta^\star+\mu u_k)
				-F_{\mathcal B_{k,r}^{(2)}}(\theta^\star)}{\mu}u_k.
		\end{align}
		The variants differ only in the minibatch used at $\theta^\star$. For $q\in\{\mathrm{match},\mathrm{mismatch}\}$, we report
		\begin{equation}
			\widehat V_q(\mu)
			=
			\frac{1}{K}\sum_{k=1}^{K}
			\frac{1}{Rd}\sum_{r=1}^{R}
			\|\hat g^q_{k,r}(\mu)-g_{F,k}(\mu)\|^2.
		\end{equation}
		
		As shown in Table~\ref{tab:controlled_noise}, $V(\mu)$ grows approximately as $1/\mu^2$ for mismatched BOLT but remains nearly constant for matched BOLT as $\mu$ decreases. This fixed-objective comparison isolates the effect of data matching.
		
		\paragraph{Direct on-policy RL objective.}
		We apply ZO finite differences directly to an on-policy expected-reward objective using 16 GSM8K prompts. The two function evaluations use independently generated on-policy rollouts, placing the estimator in the mismatched-data regime. Since the objective cannot be evaluated exactly with finite rollouts, we estimate the conditional oracle-noise variance using $K$ Gaussian perturbation directions, each with $R=32$ independent rollout pairs:
		\begin{equation}
				\widehat V_{\mathrm{RL}}(\mu)
			=
				\frac{1}{K}\sum_{k=1}^{K}
				\frac{\|u_k\|^2}{d(R-1)}
				\sum_{r=1}^{R}
				\bigl(\hat\delta_{k,r}(\mu)-\overline\delta_k(\mu)\bigr)^2.
		\end{equation}
		Here $\hat\delta_{k,r}(\mu)=[\tilde J(\theta^\star+\mu u_k,\xi_{k,r,+})-\tilde J(\theta^\star,\xi_{k,r,-})]/\mu$ denotes the scalar finite-difference coefficient for rollout pair $r$ along direction $u_k$, and $\overline\delta_k(\mu)=R^{-1}\sum_{r=1}^{R}\hat\delta_{k,r}(\mu)$ is its average across repetitions.
		
		For direct on-policy RL, Table~\ref{tab:controlled_noise} shows that $V(\mu)$ also grows approximately as $1/\mu^2$ as $\mu$ decreases. Figure~\ref{fig:oracle_noise_scaling} compares all three estimators across the full sweep and includes the predicted $1/\mu^2$ reference trend.
		
		\subsubsection{End-to-End Matched-versus-Mismatched Optimization}
		\label{app:matching_optimization}
		
		\textbf{Methods compared.}
		We compare matched- and mismatched-data variants of ZO-BOLT and ZO-GRPO using Qwen2.5-1.5B on GSM8K. The ZO-BOLT variants optimize the same fixed BOLT objective, differing only in whether paired function evaluations share a minibatch. Both ZO-GRPO variants follow the standard GRPO old-policy construction~\cite{shao2024deepseekmath}, with periodically refreshed rollouts from a frozen behavior policy. They differ only in whether the paired evaluations use independently sampled or shared rollout batches. Thus, both ZO-GRPO variants optimize the same periodically refreshed frozen-distribution GRPO surrogate rather than the true on-policy RL objective.
		
		\textbf{Common training setup.}
		All four runs train 1.09M LoRA parameters of Qwen2.5-1.5B in bf16, with rank $r=8$, scaling $\alpha=16$, zero LoRA dropout, and adapters on $q_{\mathrm{proj}}$ and $k_{\mathrm{proj}}$; the base-model parameters remain frozen. Each run uses two epochs and a seed of 42.
		
		\textbf{ZO optimization.}
		All variants use symmetric finite differences with Gaussian directions and the ZoAR~\cite{qiu2025zerothorderoptimizationsecretlysinglestep}+R-AdaZO~\cite{shu2025refining} optimizer, with a constant learning rate of $10^{-4}$. The ZO-BOLT instantiation is detailed in Alg.~\ref{alg:zobolt_full}. ZO-BOLT uses perturbation scale $\mu=10^{-3}$ and retains up to 40 historical query pairs in addition to the current pair; both ZO-GRPO controls use the historical-run values $\mu=10^{-2}$ and 30 historical pairs.
		
		\textbf{Evaluation protocol.}
		Evaluation uses the Qwen boxed-answer template, greedy decoding, a 1,024-token cap, and mathematical-equivalence grading. Historical preprocessing retains 1,317 GSM8K test examples. Checkpoint sweeps start at Step~50; Tables~\ref{tab:zobolt_matching_checkpoints} and~\ref{tab:zoogrpo_matching_checkpoints} show the separately measured pretrained baseline ($52.24\%$) as Step~0.
		
		\paragraph{ZO-BOLT control.}
		We generate eight candidate completions for each of the 7,473 GSM8K training prompts from the base Qwen2.5-1.5B model, using temperature $0.8$, top-$p$ $0.95$, at most 1,024 new tokens, and seed 42. After assigning binary boxed-correctness rewards, removing prompts with eight incorrect candidates, and truncating each retained response after its first complete boxed answer, we obtain a fixed dataset of 6,509 prompts and 52,072 completions and construct prompt-normalized Boltzmann weights with temperature $0.1$.
		
		Both variants train for 814 steps on this dataset, with prompt and completion caps of 256 and 1,024 tokens and per-device and effective batch size 128. They differ only in whether the second function evaluation reuses the first minibatch or draws another 128 examples. Table~\ref{tab:zobolt_matching_checkpoints} reports the checkpoints.
		
		\begin{table}[H]
			\caption{Checkpoint-wise GSM8K test accuracy ($\%$) for ZO-BOLT in the matched- and mismatched-data regimes. The displayed Step~0 is the separately measured pretrained baseline ($52.24\%$); the checkpoint sweeps begin at Step~50.}
			\label{tab:zobolt_matching_checkpoints}
			\centering
			\small
			\begin{tabular}{rcc@{\hspace{1.5em}}rcc}
				\toprule
				\textbf{Step} & \textbf{Matched} & \textbf{Mismatched} & \textbf{Step} & \textbf{Matched} & \textbf{Mismatched} \\
				\midrule
				0   & 52.24 & 52.24 & 450 & 60.82 & 42.60 \\
				50  & 52.85 & 49.35 & 500 & 61.43 & 45.56 \\
				100 & 53.61 & 48.52 & 550 & 63.33 & 45.25 \\
				150 & 55.88 & 46.85 & 600 & 63.48 & 44.95 \\
				200 & 54.37 & 48.75 & 650 & 63.78 & 43.58 \\
				250 & 56.72 & 45.25 & 700 & 63.55 & 40.24 \\
				300 & 56.87 & 43.20 & 750 & 62.34 & 41.53 \\
				350 & 60.14 & 41.38 & 800 & 62.41 & 38.72 \\
				400 & 61.43 & 42.29 & 814 & 62.95 & 41.76 \\
				\bottomrule
			\end{tabular}
		\end{table}
		
		\paragraph{ZO-GRPO control.}
		Both variants train directly on all 7,473 GSM8K training prompts and generate eight completions per prompt online. Each sampled rollout set contains 16 prompts and 128 completions and is processed with per-device batch size 8 and gradient accumulation 16. We use the GRPO clipped surrogate with clipping parameter $0.2$, group-normalized binary boxed rewards, no reference-policy KL penalty, and dropout disabled. For both variants, the behavior policy and rollout data are refreshed every 20 optimizer steps. Both runs use maximum prompt length 256 and 934 optimizer steps.
		
		At each rollout refresh, the mismatched-data variant independently samples two 128-completion batches from the frozen behavior policy, one for each function evaluation, whereas the matched-data variant reuses a single batch and its GRPO statistics for both evaluations. The behavior policy is synchronized with the current parameters every 20 optimizer steps, after which new rollouts are sampled. Thus, both variants optimize a periodically refreshed frozen-distribution surrogate (App.~\ref{app:rl:ppo}), differing only in whether the paired evaluations share the same data. Table~\ref{tab:zoogrpo_matching_checkpoints} reports all evaluated checkpoints.
		
		\begin{table}[H]
			\caption{Checkpoint-wise GSM8K test accuracy ($\%$) for ZO-GRPO in the matched- and mismatched-data regimes. The displayed Step~0 is the separately measured pretrained baseline ($52.24\%$); the checkpoint sweeps begin at Step~50.}
			\label{tab:zoogrpo_matching_checkpoints}
			\centering
			\small
			\begin{tabular}{rcc@{\hspace{1.5em}}rcc}
				\toprule
				\textbf{Step} & \textbf{Matched} & \textbf{Mismatched} & \textbf{Step} & \textbf{Matched} & \textbf{Mismatched} \\
				\midrule
				0   & 52.24 & 52.24 & 500 & 54.82 & 47.68 \\
				50  & 53.68 & 53.08 & 550 & 55.66 & 49.35 \\
				100 & 52.47 & 51.40 & 600 & 55.05 & 46.39 \\
				150 & 52.62 & 49.58 & 650 & 56.72 & 42.60 \\
				200 & 53.76 & 49.81 & 700 & 57.33 & 42.22 \\
				250 & 55.50 & 51.10 & 750 & 59.00 & 42.52 \\
				300 & 56.19 & 47.99 & 800 & 57.86 & 43.96 \\
				350 & 56.87 & 47.23 & 850 & 58.54 & 43.74 \\
				400 & 57.18 & 47.08 & 900 & 58.92 & 42.29 \\
				450 & 56.04 & 46.85 & 934 & 57.56 & 44.80 \\
				\bottomrule
			\end{tabular}
		\end{table}
		
		\subsection{ZO-BOLT Effectiveness}
		\label{app:zobolt_experiments}
		
		\textbf{Methods compared.}
		We compare ZO-BOLT with the pretrained baseline and first-order BOLT on the same fixed, reference-sampled, prompt-normalized Boltzmann-weighted SFT objective.
		
		\textbf{Training hyperparameters.} Both BOLT training methods use LoRA with rank $r=8$ and $\alpha=16$ applied to $q_{\text{proj}}$ and $k_{\text{proj}}$ and train in bf16 precision. First-order BOLT uses a learning rate of $5\times10^{-6}$ with cosine decay and 10\% warmup, and an effective batch size of 128. ZO-BOLT uses the ZoAR+R-AdaZO optimizer with a constant learning rate of $10^{-4}$, LoRA dropout set to 0, \texttt{zo\_eps}$=0.001$, and \texttt{num\_histories}$=40$. On GSM8K, prompt and completion lengths are capped at 256 and 1,024 tokens, respectively, for both methods. For MATH, these limits are extended to 512 and 2,048 tokens, and ZO-BOLT uses $\mu=10^{-2}$ and \texttt{num\_histories}$=30$; all other settings remain consistent with GSM8K.
		
		\textbf{Compute requirements.} All experimental runs were performed on a server equipped with two NVIDIA L40 GPUs (48GB VRAM per card). Individual runs took 5 to 20 hours, depending on the task and model size.
		
		\begin{algorithm}[t]
			\caption{ZO-BOLT with Symmetric Finite Differences, ZoAR Query Reuse, and R-AdaZO}
			\label{alg:zobolt_full}
			\begin{algorithmic}[1]
				\Require
				Fixed BOLT dataset $\mathcal D=\{(x_i,y_i,w_i)\}_{i=1}^M$ with weights $w_i$ independent of $\boldsymbol\theta$;
				initial parameters $\boldsymbol{\theta}_0\in\mathbb R^d$;
				perturbation scale $\mu>0$;
				minibatch size $B$;
				number of retained historical query pairs $N_h\geq0$;
				number of optimization steps $T$;
				learning-rate schedule $\{\eta_t\}_{t=1}^T$;
				moment coefficients $\beta_1,\beta_2\in[0,1)$;
				numerical constant $\varepsilon>0$;
				weight-decay coefficient $\lambda\geq0$;
				weight-decay mask $\mathbf{D}\in\{0,1\}^d$;
				warm-start threshold $W$

				\State Define $\ell_i(\boldsymbol\theta)=-w_i\log\pi_{\boldsymbol\theta}(y_i\mid x_i)$ and $F_{\mathcal B}(\boldsymbol\theta)=|\mathcal B|^{-1}\sum_{i\in\mathcal B}\ell_i(\boldsymbol\theta)$
				
				\State Initialize $\mathbf{m}_0\gets\mathbf 0$, $\mathbf{v}_0\gets\mathbf 0$,
				and query buffer $\mathcal Q_0\gets\varnothing$

				\For{$t=1,\ldots,T$}
				
				\State Sample minibatch $\mathcal B_t\subseteq\{1,\ldots,M\}$ of size $B$
				\State Sample seed $r_t$ and generate $\mathbf{u}_t\sim\mathcal N(0,I_d)$
				
				\State Query:
				$y_t^+\gets F_{\mathcal B_t}(\boldsymbol{\theta}_{t-1}+\mu \mathbf{u}_t)$,
				$y_t^-\gets F_{\mathcal B_t}(\boldsymbol{\theta}_{t-1}-\mu \mathbf{u}_t)$
				
				\State Update query buffer:
				$\mathcal Q_t\gets
				\operatorname{KeepMostRecent}_{N_h+1}
				(\mathcal Q_{t-1}\cup\{(r_t,y_t^+,y_t^-)\})$
				
				\State Set $M_t\gets|\mathcal Q_t|$ and recover each $\mathbf{u}_s$ from stored seed $r_s$
				
				\State Compute query-reused ZO estimator:
				$\mathbf{g}_t\gets
				\frac1{M_t}
				\sum_{(r_s,y_s^+,y_s^-)\in\mathcal Q_t}
				\frac{y_s^+-y_s^-}{2\mu}\mathbf{u}_s$
				
				\State Apply coupled weight decay:
				$\bar{\mathbf{g}}_t\gets \mathbf{g}_t+\lambda \mathbf{D}\odot\boldsymbol{\theta}_{t-1}$
				
				\State Update moments:
				$\mathbf{m}_t\gets\beta_1\mathbf{m}_{t-1}+(1-\beta_1)\bar{\mathbf{g}}_t$
				
				\State Update refined second moment:
				$\mathbf{v}_t\gets\beta_2\mathbf{v}_{t-1}+(1-\beta_2)\mathbf{m}_t^2$
				
				\If{$t<W$}
				\State $\widetilde{\mathbf{v}}_t\gets\mathbf 1$
				\Else
				\State $\widetilde{\mathbf{v}}_t\gets\mathbf{v}_t$
				\EndIf
				
				\State Update parameters:
				$\boldsymbol{\theta}_t\gets
				\boldsymbol{\theta}_{t-1}
				-
				\eta_t
				\frac{\mathbf{m}_t}{\sqrt{\widetilde{\mathbf{v}}_t}+\varepsilon}$
				
				\EndFor
				
				\State \Return $\boldsymbol{\theta}_T$
				
			\end{algorithmic}
		\end{algorithm}

		\paragraph{Implementation details.}
		The fixed dataset and weights are constructed before optimization following BOLT~\cite{shu2026referencesampledboltzmannprojectionklregularized}; Algorithm~\ref{alg:zobolt_full} uses only their resulting weighted-SFT objective. At each optimization step, symmetric finite differences evaluate an antithetic query pair on the same minibatch, and ZoAR retains at most $N_h$ historical pairs. Each pair stores only its random seed and two scalar function values, with the direction regenerated deterministically during updates. The unperturbed loss is used only for logging.
		
		We use $\beta_1=0.9$, $\beta_2=0.99$, and $\varepsilon=10^{-8}$ without bias correction. The adaptive denominator is set to one for the first $W$ updates, after which the accumulated second moment $\mathbf v_t$ is used. The gradient-accumulation factor is one.

		\subsubsection{ZoAR+R-AdaZO and the forward finite-difference estimator}
		\label{app:optimizer_theory}
		Algorithm~\ref{alg:zobolt} outlines ZO-BOLT, while Algorithm~\ref{alg:zobolt_full} specifies its implementation details. We analyze the oracle-noise mechanism rather than establishing variance or convergence guarantees for the complete history-reused, adaptively preconditioned optimizer.
	
		\paragraph{Query reuse and R-AdaZO.}
		In this analysis, we return to the generic oracle notation $\widetilde F(\theta;\xi)$ to cover both data regimes. At iteration $t$, ZoAR maintains a buffer $\mathcal Q_t$ containing the current query pair and at most $N_h$ historical pairs. With retained indices $\mathcal I_t$ and $M_t=|\mathcal Q_t|$, the reused gradient estimate is
		\begin{equation}
		\label{eq:app_query_reused_estimator}
		g_t=\frac1{M_t}\sum_{s\in\mathcal I_t}\frac{y_s^+-y_s^-}{2\mu}\,u_s,
		\qquad
		y_s^\pm=\widetilde F(\theta_{s-1}\pm\mu u_s,\xi_{s,\pm}).
		\end{equation}
		Each historical pair is reused at its original iterate without additional function evaluations. R-AdaZO then performs first-moment smoothing and adaptive preconditioning (Algorithm~\ref{alg:zobolt_full}), accumulating the squared first-moment estimate $m_t^2$ rather than $g_t^2$.

		\paragraph{Relation to the forward estimator.}
		The analyzed forward estimator and implemented symmetric estimator are
		\begin{equation}
		\label{eq:app_estimator_forms}
		\begin{aligned}
		\hat g_{\mathrm{fwd}}&=\frac{\widetilde F(\theta+\mu u,\xi_+)-\widetilde F(\theta,\xi_-)}{\mu}\,u,\\
		\hat g_{\mathrm{sym}}&=\frac{\widetilde F(\theta+\mu u,\xi_+)-\widetilde F(\theta-\mu u,\xi_-)}{2\mu}\,u.
		\end{aligned}
		\end{equation}
		For the noise-free counterpart $g_F^{\mathrm{sym}}(\theta;u):=[F(\theta+\mu u)-F(\theta-\mu u)]u/(2\mu)$, with $e_\pm:=e(\theta\pm\mu u,\xi_\pm)$, we have
		\begin{equation}
		\label{eq:app_symmetric_decomposition}
		\hat g_{\mathrm{sym}}-g_F^{\mathrm{sym}}=\frac{e_+-e_-}{2\mu}\,u,
		\end{equation}
		and consequently,
		\begin{equation}
		\label{eq:app_symmetric_noise_identity}
		\mathbb E[\|\hat g_{\mathrm{sym}}-g_F^{\mathrm{sym}}\|^2\mid\theta]
		=\frac1{4\mu^2}\mathbb E[(e_+-e_-)^2\|u\|^2\mid\theta].
		\end{equation}
		Thus, both estimators share the same paired oracle-noise mechanism, with the symmetric estimator changing only the scaling constant, not the $1/\mu^2$ amplification under independent oracle noise.
		
		\paragraph{History reuse and adaptive updates.}
		Let $e_{s,\pm}=e(\theta_{s-1}\pm\mu u_s,\xi_{s,\pm})$. The aggregated oracle-noise component satisfies
		\begin{equation}
		\label{eq:app_reused_noise_decomposition}
		g_t-\frac1{M_t}\sum_{s\in\mathcal I_t}g_F^{\mathrm{sym}}(\theta_{s-1};u_s)
		=\frac1{M_t}\sum_{s\in\mathcal I_t}\frac{e_{s,+}-e_{s,-}}{2\mu}\,u_s.
		\end{equation}
		Under conditionally independent pairwise noise, history averaging reduces the variance coefficient by approximately $1/M_t$ but does not eliminate the $1/\mu^2$ dependence of mismatched evaluations. R-AdaZO leaves the pairwise oracle-noise decomposition unchanged, although its adaptive updates may affect noise propagation during optimization.
		
		\subsection{Memory Efficiency}
		\label{app:memory_experiments}
		
		We measure peak GPU memory on NVIDIA L40 GPUs for Qwen2.5-1.5B and Qwen2.5-7B. All compared methods use LoRA with $r=8$ and $\alpha=16$, batch size 2, maximum sequence length 1,024, and bf16 precision. Table~\ref{tab:memory_combined} reports both the measured peak and its absolute overhead over inference-only execution. These measurements are intended to verify that the forward-only ZO implementations retain the standard near-inference memory profile; they are not presented as a memory-saving contribution specific to ZO-BOLT.

		
		\newpage
		\input{checklist.tex}

	\end{document}

%% file: checklist.tex
\section*{NeurIPS Paper Checklist}

\begin{enumerate}

\item {\bf Claims}
    \item[] Question: Do the main claims made in the abstract and introduction accurately reflect the paper's contributions and scope?
    \item[] Answer: \answerYes{} 
    \item[] Justification: The main claims made in the abstract and introduction are verified theoretically and empirically in Section~\ref{sec:ZOOnoise} and Section~\ref{sec:experiments}.

\item {\bf Limitations}
    \item[] Question: Does the paper discuss the limitations of the work performed by the authors?
    \item[] Answer: \answerYes{} 
    \item[] Justification: We mention our limitations in Section~\ref{sec:conclusion}.

\item {\bf Theory assumptions and proofs}
    \item[] Question: For each theoretical result, does the paper provide the full set of assumptions and a complete (and correct) proof?
    \item[] Answer: \answerYes{} 
    \item[] Justification: We provide detailed proofs in Appendix~\ref{sec:appendix_proofs}.

    \item {\bf Experimental result reproducibility}
    \item[] Question: Does the paper fully disclose all the information needed to reproduce the main experimental results of the paper to the extent that it affects the main claims and/or conclusions of the paper (regardless of whether the code and data are provided or not)?
    \item[] Answer: \answerYes{} 
    \item[] Justification: We provide the experimental setup and relevant parameters required for reproduction in the Appendix~\ref{app:exp_setup}.

\item {\bf Open access to data and code}
    \item[] Question: Does the paper provide open access to the data and code, with sufficient instructions to faithfully reproduce the main experimental results, as described in supplemental material?
    \item[] Answer: \answerYes{} 
    \item[] Justification: We will provide data, code, and instructions for the final paper.

\item {\bf Experimental setting/details}
    \item[] Question: Does the paper specify all the training and test details (e.g., data splits, hyperparameters, how they were chosen, type of optimizer) necessary to understand the results?
    \item[] Answer: \answerYes{} 
    \item[] Justification: We provide the experimental setup and relevant parameters required to understand the results in the Appendix~\ref{app:exp_setup}.

\item {\bf Experiment statistical significance}
    \item[] Question: Does the paper report error bars suitably and correctly defined or other appropriate information about the statistical significance of the experiments?
    \item[] Answer: \answerNo{} 
    \item[] Justification: Due to computational resource constraints, error bars are not reported, as doing so would be expensive.

\item {\bf Experiments compute resources}
    \item[] Question: For each experiment, does the paper provide sufficient information on the computer resources (type of compute workers, memory, time of execution) needed to reproduce the experiments?
    \item[] Answer: \answerYes{} 
    \item[] Justification: We provide information about the computer resources used in Section~\ref{app:exp_setup}.
    
\item {\bf Code of ethics}
    \item[] Question: Does the research conducted in the paper conform, in every respect, with the NeurIPS Code of Ethics \url{https://neurips.cc/public/EthicsGuidelines}?
    \item[] Answer: \answerYes{} 
    \item[] Justification: We have read and understood the code of ethics and have made every effort to adhere to it.

\item {\bf Broader impacts}
    \item[] Question: Does the paper discuss both potential positive societal impacts and negative societal impacts of the work performed?
    \item[] Answer: \answerNA{} 
    \item[] Justification: The work is foundational research on optimization algorithms for LLM fine-tuning. It focuses on theoretical analysis and controlled experiments, without proposing a deployment or application that would have direct societal impacts. No specific positive or negative societal consequences arise from the methods themselves.
    
\item {\bf Safeguards}
    \item[] Question: Does the paper describe safeguards that have been put in place for responsible release of data or models that have a high risk for misuse (e.g., pre-trained language models, image generators, or scraped datasets)?
    \item[] Answer: \answerNA{} 
    \item[] Justification: The paper does not release any new models or datasets. Our work uses standard, publicly available models (Qwen2.5) and datasets(GSM8K, MATH), and does not pose additional risk of misuse beyond their existing public availability.

\item {\bf Licenses for existing assets}
    \item[] Question: Are the creators or original owners of assets (e.g., code, data, models), used in the paper, properly credited and are the license and terms of use explicitly mentioned and properly respected?
    \item[] Answer: \answerYes{} 
    \item[] Justification: We use publicly available datasets (GSM8K, MATH) and pretrained models (Qwen2.5), all of which are properly cited in the paper. We adhere to the licenses and terms of use as specified by their original creators.

\item {\bf New assets}
    \item[] Question: Are new assets introduced in the paper well documented and is the documentation provided alongside the assets?
    \item[] Answer: \answerYes{} 
    \item[] Justification: The paper will release the experimental code used for reproducing the results in the final version. The code is documented with instructions for setup, dependencies, and running experiments. 

\item {\bf Crowdsourcing and research with human subjects}
    \item[] Question: For crowdsourcing experiments and research with human subjects, does the paper include the full text of instructions given to participants and screenshots, if applicable, as well as details about compensation (if any)? 
    \item[] Answer: \answerNA{}
    \item[] Justification: The paper does not involve crowdsourcing or research with human subjects.

\item {\bf Institutional review board (IRB) approvals or equivalent for research with human subjects}
    \item[] Question: Does the paper describe potential risks incurred by study participants, whether such risks were disclosed to the subjects, and whether Institutional Review Board (IRB) approvals (or an equivalent approval/review based on the requirements of your country or institution) were obtained?
    \item[] Answer: \answerNA{}
    \item[] Justification: The paper does not involve research with human subjects, so no IRB or equivalent approvals were required.

\item {\bf Declaration of LLM usage}
    \item[] Question: Does the paper describe the usage of LLMs if it is an important, original, or non-standard component of the core methods in this research? Note that if the LLM is used only for writing, editing, or formatting purposes and does \emph{not} impact the core methodology, scientific rigor, or originality of the research, declaration is not required.
    \item[] Answer: \answerNA{} 
    \item[] Justification: We used an LLM solely for editing assistance.

\end{enumerate}